\documentclass[letterpaper]{article}
\usepackage{aaai}
\usepackage{times}
\usepackage{helvet}
\usepackage{courier}
\usepackage{color}
\usepackage{todonotes}
\usepackage{graphicx, subfigure}
\usepackage{amssymb}
\usepackage{epstopdf}
\usepackage{amsmath}
\usepackage{enumerate}
\usepackage{amsthm}
\usepackage{undertilde}
\usepackage{lipsum}

\newcommand{\citet}[1]{\citeauthor{#1} (\citeyear{#1})}

\newcommand\blfootnote[1]{%
  \begingroup
  \renewcommand\thefootnote{}\footnote{#1}%
  \addtocounter{footnote}{-1}%
  \endgroup
}

\newcommand{\ux}{\underline{x}}
\newcommand{\ox}{\overline{x}}
\newcommand{\un}{\underline{n}}
\newcommand{\on}{\overline{n}}
\newcommand{\af}{\alpha}
\newcommand{\bt}{\beta}
\newcommand{\splus}{\hspace{-0.04cm}+\hspace{-0.04cm}}
\newcommand{\sminus}{\hspace{-0.04cm}-\hspace{-0.04cm}}
\newcommand{\urho}{\underline{\rho}}
\newcommand{\orho}{\overline{\rho}}
\newcommand{\urhof}{\frac{\underline{\rho}}{2}}
\newcommand{\orhof}{\frac{\overline{\rho}}{2}}
\newcommand{\urhopf}{\frac{\underline{\rho}'}{2}}
\newcommand{\orhopf}{\frac{\overline{\rho}'}{2}}

\newtheorem{thm}{Theorem}
\newtheorem{lem}[thm]{Lemma}
\newtheorem{rem}[thm]{Remark}
\begin{document}
\title{Proximal Operators for Multi-Agent Path Planning}
%\author{Jos\'e Bento$^\dagger$, Nate Derbinsky, Charles Mathy, Jonathan S. Yedidia\\
%\hspace{1cm}\\
%\emph{jose.bento@bc.edu, derbinskyn@wit.edu, cmathy@disneyresearch.com, yedidia@disneyresearch.com}
%}
\author{Jos\'e Bento$^\dagger$\\
Boston College\\
jose.bento@bc.edu\\
\And
Nate Derbinsky\\
Wentworth Institute of Technology\\
derbinskyn@wit.edu\\
\And
Charles Mathy\\
Disney Research Boston\\
cmathy@disneyresearch.com\\
\And
Jonathan S. Yedidia\\
Disney Research Boston\\
yedidia@disneyresearch.com\\
}
\maketitle

%
%************************************
%
\begin{abstract}
\begin{quote}
We address the problem of planning collision-free 
paths for multiple agents using optimization methods
known as \emph{proximal algorithms}. Recently this approach was explored in 
\citet{bento2013message}, which
demonstrated its
ease of parallelization and decentralization, the speed
with which the algorithms generate good quality solutions, and
its ability to incorporate different \emph{proximal operators},
each ensuring that paths satisfy a desired property.
Unfortunately, the operators derived only apply to paths in 2D
and require that any intermediate waypoints 
we might want agents to follow be preassigned to specific agents,
limiting their range of applicability.
In this paper we resolve these limitations.
We introduce new operators to deal with agents moving
in arbitrary dimensions that are faster to compute than their
2D predecessors and we introduce \emph{landmarks},
space-time positions that are automatically assigned
to the set of agents under different optimality criteria.
Finally, we report the performance of the new operators
in several numerical experiments\blfootnote{$\dagger$A special thanks to Lucas Foguth for valuable discussions.}.
\end{quote}
\end{abstract}
%
%************************************
%

%include all the files
\vspace{-6mm}
\section{Introduction} \label{sec:intro}

% something that we want that is not easy to do
In this paper we provide a novel set of algorithmic building blocks (proximal operators)
to plan paths for a system of multiple independent robots that
need to move optimally across a set of locations and avoid collisions with
obstacles and each other. This problem is crucial in applications involving
automated storage, exploration and surveillance.

Even if each robot has few degrees of freedom, the joint system is complex and 
this problem is hard to solve \cite{reif79,hopcroft84}.
%
% literature review
We can divide existing algorithms for this problem
into \emph{global planners}, if they find collision-free beginning-to-end paths connecting
two desired configurations, or \emph{local planners}, if they find short collision-free paths 
that move the system only a bit closer to the final configuration.

We briefly review two of the most rigorous approaches.
%We now mention two: the approach with most rigourous optimality guarantees
%and the approach closest to our work.
%
% sampling methods
Random sampling methods, first introduced in \cite{kavraki94,kavraki96},
are applicable to global planning and explore the space of possible robot configurations
with discrete structures.
The rapidly-exploring random tree algorithm (RRT; \citeauthor{lavalle01} \citeyear{lavalle01}),
is guaranteed to asymptotically find feasible solutions with high-probabilty 
while the RRT* algorithm \cite{karaman2010incremental} asymptotically finds the optimal solution.
However, their convergence rate degrades as the
dimension of the joint configuration space increases, as when considering multiple robots,
and they cannot easily find solutions where robots move in tight spaces.
In addition, even approximately solving
some simple problems requires many samples, e.g., approximating
a shortest path solution for a single robot required to move between two points
with no obstacles \cite[see Fig. 1]{karaman2010incremental}.
These methods explore a continuous space using discrete
structures and are different from methods that only consider
agents that move on a graph with no concern
about their volume or dynamics, e.g. \cite{standley2011complete,sharon2013increasing}.

% optimization based methods
An optimization-based approach has been used by several authors, including
\citet{mellinger12}, who formulate
global planning as a mixed-integer quadratic problem and, for up to four robots, solve it
using branch and bound techniques. Sequential convex 
programming was used in \cite{augugliaro12} 
to efficiently obtain good local optima for global planning up to twelve robots.
State-of-the-art optimization-based algorithms for local planning typically
have real-time performance and are based on the \emph{velocity-obstacle}
(VO) idea of \cite{fiorini98},
which greedily plans paths only a few seconds into the future and then re-plans.
These methods scale to hundreds of robots.
Unlike sampling algorithms, optimization-based methods
easily find solutions where robots move tightly together and solve
simple problems very fast. However, they do not perform as well in problems
involving robots in complex mazes.

% a very quick review of the work done by bento2013 on robot planning
Our work builds on the work of \citet{bento2013message}, which formulates
multi-agent path planning as a large non-convex optimization problem and uses
\emph{proximal algorithms} to solve it.
More specifically, the authors use
the Alternating Direction Method of Multipliers (ADMM) and the
variant introduced in \citet{nate13} called the Three Weight Algorithm (TWA).
% short explanation of ADMM and TWA
These are iterative algorithms that do not access the objective function
directly but indirectly through multiple (simple) algorithmic blocks called
\emph{proximal operators}, one per function-term in the objective. At each
iteration these operators can be applied
independently and so both the TWA and the ADMM are easily parallelized.
A brief explanation of this optimization formulation is
given in Section \ref{sec:backG}.
A self-contained explanation about proximal algorithms is
found in \citet{parikh2013proximal} and a good review on the ADMM is \citet{boyd2010}.
% there is no guarantee of convergence but the results are good
In general the ADMM and the TWA are not guaranteed to converge for non-convex problems.
There is some work on solving non-convex problems using
proximal algorithms with guarantees (see \citeauthor{udell2014bounding} \citeyear{udell2014bounding} and references in
\citeauthor{parikh2013proximal} \citeyear{parikh2013proximal}) but the settings considered are not
applicable to the optimization problem at hand.
Nonetheless, the empirical results in \citet{bento2013message} are very satisfactory.
For global planning, their algorithm 
scales to many more robots than other optimization based methods
and finds better solutions than VO-based methods.
Their method also can be implemented in the (useful) form of a decentralized
message-passing
scheme and new proximal operators can be easily added or removed to account for different
aspects of planning, such as, richer dynamic and obstacle constraints.

%limitation of previous work
The main contributions of \citet{bento2013message} are
the proximal operators that enforce no robot-robot collisions and no robot-wall collisions.
These operators involve solving a non-trivial problem with an infinite number
of constraints and a finite number of variables, also known
as a \emph{semi-infinite programming} problem (SIP). The
authors solve this SIP problem only for robots moving in 2D by means
of a mechanical analogy, which unfortunately excludes applications in 3D
such as those involving fleets of unmanned aerial vehicles (UAVs) or
autonomous underwater vehicles (AUVs).
Another limitation of their work is that it does not allow
robots to automatically select waypoint positions from
a set of reference positions. This is required, for example, in problems involving robots in
formations \cite{bahceci2003review}. In \citet{bento2013message},
any reference position
must be pre-assigned to a specific robot.

In this paper we propose
a solution to these limitations. Our contributions are
%and our contribution
(i) we rigorously prove that the SIP problem involved in collision proximal operators can be
reduced to solving a single-constraint non-convex problem that we solve explicitly
in arbitrary dimensions and numerically show our novel approach is substantially faster for 2D than
\citet{bento2013message} and
(ii) we derive new proximal operators that automatically
assign agents to a subset of reference positions and penalize non-optimal
assignments.
Our contributions have an impact beyond path planning problems. 
Other applications in robotics, computer vision or CAD
that can be tackled via large optimization problems
involving collision constraints
or the optimal assignment of objects to positions
(e.g. \citeauthor{kuffner2002self} \citeyear{kuffner2002self}; \citeauthor{witkin1988spacetime} \citeyear{witkin1988spacetime}; \citeauthor{andreev2001metal} \citeyear{andreev2001metal})
might benefit from our new building blocks (cf. Section \ref{sec:concl}).

%comment on literature about SIP
While there is an extensive literature on how to solve SIP problems (see \citet{stein2012solve} for a good review),
as far as we know, previous methods are either too general and, 
when applied to our problem, computationally more expensive than our approach, or too
restrictive and thus not applicable.

% small clarification about the focus of our paper
Finally, we clarify that our paper is not so much about showing the merits
of the framework used in \citeauthor{bento2013message} (\citeyear{bento2013message}; a point already made),
as it is about overcoming unsolved critical limitations.
However, our
numerical results and supplementary video do confirm that the framework
produces very good results, although
there are no guarantees that the method avoids
local minima.

\vspace{-0.2cm}
\section{Background} \label{sec:backG}
%
%\vspace{-0.0cm}
%
% summarize the section
Here we review the formulation
of \citet{bento2013message} of path planning as an optimization
problem, explain what proximal operators are, and
explain their connection to solving this optimization problem.

% describe the variables
We have $p$ spherical agents in $\mathbb{R}^d$ of radius $\{r_i\}^p_{i = 1}$.
Our objective is
to find collision-free paths $\{x_i(t)\}_{i \in [p],t \in [0,T]}$ for all agents
between their specified initial positions $\{x^{\text{init}}_i\}^p_{i  = 1}$ at time $0$ and
specified final positions $\{x^{\text{final}}_i\}^p_{i  = 1}$ at time $T$. In the simplest
case, we divide time in intervals of equal length
and the path $\{x_i(t)\}^T_{t = 0}$ of agent
$i \hspace{-1mm}\in\hspace{-1mm} [p]$
is parametrized by a set of \emph{break-points} $\{x_i(s)\}^\eta_{s = 0}$ such that
$x_i(t) \hspace{-1mm}=\hspace{-1mm} x_i(s)$ for $t \hspace{-1mm}=\hspace{-1mm} s T/\eta$ and all $s$.
Between break-points agents have zero-acceleration. We discuss the practical impact of this assumption in Appendix \ref{app:path_smoothness}.
%In Appendix \ref{app:path_smoothness} we discuss the practical impact
%of the piece-wise linear assumption.

%describe the optimization problem
We express global path planning as an
optimization problem with an objective function that is a large
sum of simple cost functions. Each function
accounts for a different aspect of the problem. Using similar notation to \citet{bento2013message}, we need to minimize the objective function
{\small
\begin{align}\label{eq:global_path_opt_problem}
%& \hspace{-0.05cm} \text{minimize} 
&\sum_i \hspace{-0.05cm}  f^{pos}(x_i(0),x^{init}_i) \hspace{-0.05cm}
+\sum_i  f^{pos}(x_i(\eta),x^{final}_i)
\hspace{-0.05cm}+  \hspace{-0.2cm} \sum_{i>j,s} \\
&f^{\text{coll}}_{i,j}(x_i(s),x_i(s\hspace{-0.1cm}+\hspace{-0.1cm}1),x_j(s),x_j(s\hspace{-0.1cm}+\hspace{-0.1cm}1)) + \hspace{-0.1cm} \sum_{i,s} f^{\text{vel}}_i(x_i(s),x_i(s\hspace{-0.1cm}+\hspace{-0.1cm}1)) \nonumber \\
&+ \hspace{-0.15cm}\sum_{\mathcal{W},i,s} f^{\text{wall}}_{\mathcal{W}} (x_i(s),x_i(s\hspace{-0.1cm}+\hspace{-0.1cm}1))
+ \hspace{-0.1cm}\sum_{i,s} f^{\text{dir}}_i(x_i(s),x_i(s\hspace{-0.1cm}+\hspace{-0.1cm}1),x_i(s\hspace{-0.1cm}+\hspace{-0.1cm}2)) 
\hspace{-0.05cm} \nonumber. 
\end{align}
}

\vspace{-3mm}
The function $f^{\text{coll}}$ prevents agent-agent collisions: it is zero if
$\|\af x_i(s) \splus (1\sminus\af) x_i(s\splus1) \sminus (\af x_j(s) \splus(1\sminus\af) x_j(s\splus1) )\| \geq r_i \splus r_j$ for all $\af \in [0,1]$
and infinity otherwise. The $f^{\text{wall}}$ function prevents agents from colliding with obstacles: it is zero if
$\|\af x_i(s) \splus (1\sminus \af) x_i(s\splus1) \sminus y\| \geq r_i$ for all $\af \in [0,1], y \in \mathcal{W}$,
where $\mathcal{W}$ is a set of points defining an obstacle, and is infinity
otherwise. In \citet{bento2013message}, $\mathcal{W}$ is a line between two points
$x_L$ and $x_R$ in the plane and the summation $\sum_{\mathcal{W}}$ is across a set of obstacles.
The functions $f^{\text{vel}}$ and $f^{\text{dir}}$ impose restrictions on the velocities and direction changes of paths.
%We can have, for example, $f^{\text{vel}}_1(x_1(s),x_1(s\splus1)) = 0.1 \| x_1(s\splus1) \sminus x_1(s)\|^2$ 
%and $f^{\text{dir}}_3(x_3(s),x_3(s\splus1),x_3(s\splus2)) = 0.03 \| x_3(s\splus2) \sminus 2x_3(s\splus1) \splus x_3(s)\|^2$.
The function $f^{\text{pos}}$ imposes boundary conditions: it is zero if $x_i(0) = x^{\text{init}}_i$
(or if $x_i(\eta) = x^{\text{final}}_i$) and infinity otherwise.
%
%Not all of the terms
%in \eqref{eq:global_path_opt_problem}
%need to be always included and we can included new terms to account for other
%aspects of planning. In Section \ref{sec:landmark} we include new terms to account for automatic
%assignment of agents to positions in space-time.
%
The authors also re-implement
the local path planning method of \citet{alonsomora13icra}, based on \emph{velocity obstacles} \cite{fiorini98}, by solving an optimization
algorithm similar to \eqref{eq:global_path_opt_problem}.

%we now quickly explain how ADMM or TWA solves this problem in a parallel fashion
\citet{bento2013message} solve \eqref{eq:global_path_opt_problem}
using the TWA, a variation of the ADMM.
The ADMM is an iterative algorithm that minimizes
objectives that are a sum of many different functions.
The ADMM 
is guaranteed to solve convex problems, but, empirically,
the ADMM (and the TWA)
can find good feasible solutions for large non-convex problems
\cite{nate13,bento2013message}.

Loosely speaking, the ADMM proceeds by passing
messages back and forth between two types of blocks:
\emph{proximal operators} and \emph{consensus operators}.
First, each function in the objective is queried
separately by its associated proximal operator
to estimate the optimal
value of the variables the function depends on. 
For example, the proximal operator associated with $f^{\text{coll}}_{1,2}$
produces estimates for optimal value of
$x_1(s),x_2(s),x_1(s+1)$ and $x_2(s+1)$.
These estimates are then sent to the consensus operators.
Second, a consensus value for each variable
is produced by its associated consensus operator
by combining all the received different estimates
for the values of the variable that the proximal operators
produced. For example,
the proximal operators associated with 
$f^{\text{coll}}_{1,2}$ and $f^{\text{vel}}_{1}$
give two different estimates for the optimal value of $x_1(s)$
and the consensus operator associated with $x_1(s)$ needs to
combine them into a single estimate.
The consensus estimates produced by the consensus
operators are then communicated back to and used by the
proximal operators to produce new estimates, 
and the cycle is repeated until convergence. See Appendix
\ref{app:ADMM_TWA_blocks}
for an illustration of the blocks that solve
a problem for two agents.

It is important to be more specific here. Consider a function $f(x)$ in the objective.
From the consensus value for its variables $x$,
the corresponding consensus nodes form consensus messages $n$ that
are sent to
the proximal operator associated with $f$. The proximal operator then
estimates the optimal value for $x$ as a tradeoff between
a solution that is close
to the minimizer of $f$ and one that is close to the consensus information
in $n$ \cite{parikh2013proximal},
\vspace{-1mm}
{
\begin{equation} \label{eq:general_proximal_operator}
x \in \arg \min_{x'} f(x')  + \frac{\rho}{2} \|x' - n \|^2,
\end{equation}
}
\vspace{-1mm}
where we use $\in$ instead of $=$ to indicate that, for a non-convex function $f$, the operator might be one-to-many, in which case some extra tie-breaking rule needs to be implemented. The variable $\rho$ is a free parameter of the ADMM that controls this tradeoff and whose
value affects its performance. In the TWA the performance is
improved by dynamically assigning to the $\rho$'s of
the different proximal operators values in $\{0,const.,\infty \}$
(cf. Appendix \ref{app:TWA_implementation}).

We emphasize that the implementation
of these proximal operators is the crucial inner-loop step of the ADMM/TWA.
For example, when $f=f^{\text{vel}}$ takes a quadratic (kinetic energy) form, 
the operator \eqref{eq:general_proximal_operator}
has a simple closed-form expression. However, for
$f = f^{\text{coll}}$ or $f = f^{\text{wall}}$ the operator involves solving
a SIP problem. In Section \ref{sec:nocoll} we explain how to
compute these operators more efficiently and in a more general setting
than in \citet{bento2013message}. 
\vspace{-1mm}
\section{No-collision proximal operator} \label{sec:nocoll}
\vspace{1mm}

Here we study the proximal operator associated with the function
$f^{\text{coll}}$ that ensures
there is no collision between two agents of radius $r$ and $r'$
that move between two consecutive break-points. We
distinguish the variables associated to the two agents using '
and distinguish the variables associated to the two break-points using
${}_{-}$ and $\overline{\color{white} \text{a}}$, respectively. 
For concreteness, just imagine, for example, that
$\ux = x_1(0), \ux' = x_2(0), \ox = x_1(1)$ and $\ox' = x_2(1)$ and think
of $\un,\un',\on'$ and $\on'$ as the associated received consensus messages.
Following \eqref{eq:general_proximal_operator}, the 
operator associated to $f^{\text{coll}}$ outputs the
minimizer of
\vspace{-4mm}
%
% minimization problem for the collision-collision minimizer
%
{\small
\begin{gather} \label{eq:coll_prox_op}
\begin{aligned}
&{\min_{\ux,\ux',\ox,\ox'}} \urhof \| \ux \sminus \un \|^2 \splus \orhof \| \ox \sminus \on \|^2 \splus \urhopf \| \ux' \sminus \un' \|^2 \splus \orhopf \| \ox' \sminus \on' \|^2 \\
&  \text{ s.t. }  
 \| \af (\ux - \ux') \splus (1 \sminus \af) (\ox \sminus \ox') \|  \geq r \splus r', {\text{\bf for all } \af \hspace{-1mm} \in \hspace{-1mm} [0,1]}.
\end{aligned}\raisetag{4mm}
\end{gather}
}
%
%
% summary of contribution of this section
%
Our most important contribution here is an efficient procedure to solve
the above semi-infinite programming problem for agents in arbitrary
dimensions by reducing it to a max-min problem. Concretely, 
Theorem \ref{th:agent_agent_coll_max_min} below
shows that \eqref{eq:coll_prox_op} is essentially equivalent
to the `most costly' of the problems in the following family
of single-constraint problems parametrized by $\af$,
\vspace{-2mm}
%
% simplified optimization problem
%
{\small
\begin{gather} \label{eq:coll_prox_op_single_alpha}
\begin{aligned}
&{\min_{\ux,\ux',\ox,\ox'}} 
\urhof \| \ux \sminus \un \|^2 \splus \orhof \| \ox \sminus \on \|^2 \splus \urhopf \| \ux' \sminus \un' \|^2 \splus \orhopf \| \ox' \sminus \on' \|^2 \\
& \;\;\; \text{s.t. } \| \af (\ux \sminus \ux') \splus (1 \sminus \af) (\ox \sminus \ox') \|  \geq r \splus r'.
\end{aligned}\raisetag{4mm}
\end{gather}
}
Since problem \eqref{eq:coll_prox_op_single_alpha} has a simple closed-form solution, we can
solve \eqref{eq:coll_prox_op} faster than
in \citet{bento2013message} for 2D objects. We support this claim with
numerical results in Section \ref{sec:num}. In the supplementary video we use our new
operator to do planning in 3D and, for illustration purposes, in 4D.
%
% we describe the main practical theorem
%
\begin{thm} \label{th:agent_agent_coll_max_min}
%If $r + r' \leq \min_{\af \in [0,1]} \| \af (\un - \un') + (1 - \af) (\on - \on')\|$ then
%problem \eqref{} and problem \eqref{} have the same unique solution for every $\af$.
%If $r + r' > \min_{\af \in [0,1]} \| \af (\un - \un') + (1 - \af) (\on - \on')\|$ and
If $\| \af (\un - \un') + (1 - \af) (\on - \on') \| \neq 0$, then
\eqref{eq:coll_prox_op_single_alpha} has a unique minimizer, $x^*(\af)$,
and if this condition holds for $\af = \af^* \in \arg \max_{\af' \in [0,1]}  h(\af')$, where 
$2h^2(\af)$ is the minimum value of \eqref{eq:coll_prox_op_single_alpha}, 
then $x^*(\af^*)$ is also a minimizer of \eqref{eq:coll_prox_op}.
In addition, if $\| \af (\un - \un') + (1 - \af) (\on - \on') \| \neq 0$, then
%
% we state the closed form expression for h(a)
%
{\small
\begin{equation} \label{eq:expression_for_h}
h(\af) = \max \left\{0, \;\frac{(r+r') - \| \af \Delta \un + (1-\af) \Delta \on \|}{\sqrt{\af^2 / \utilde{\rho} + (1-\af)^2 / \tilde{\rho}}}\right\},
\end{equation}
}
and the unique
minimizer of \eqref{eq:coll_prox_op_single_alpha} is
%
% we state the expressions to compute the outgoing X
%
{\small
\begin{align}
\ux^* &=  \un - \gamma \urho ({\af}^2 \Delta \un + \af(1-\af) \Delta \on ), \label{eq:x_out_coll_1}\\
\ux'^* &= \un' + \gamma \urho' ({\af}^2 \Delta \un + \af(1-\af) \Delta \on ),\\
\ox^* &=  \on - \gamma \orho ((1- \af )\af \Delta \un + (1-\af)^2 \Delta \on),\\ 
\ox'^* &=  \on' + \gamma \orho' ((1- \af )\af \Delta \un + (1-\af)^2 \Delta \on) \label{eq:x_out_coll_4},
\end{align}
}
where $\utilde{\rho} = (\urho^{-1} + {\urho'}^{-1})^{-1}$, $\tilde{\rho} = (\orho^{-1} + {\orho'}^{-1})^{-1}$, $\gamma\hspace{-0mm}=\hspace{-0mm}\frac{2 \lambda}{1 + 2 \lambda ({\af}^2 / \utilde{\rho} + (1-\af)^2 / \tilde{\rho})}$, $\lambda \hspace{-0mm}=\hspace{-0mm}- \frac{ h(\af)}{ 2(r + r') \sqrt{{\af}^2 / \utilde{\rho} + (1-\af)^2 / \tilde{\rho}}}$, $\Delta \un = \un - \un'$ and $\Delta \on = \on - \on'$. \end{thm}
%
% here we make the remark that, in general, the above theorem will give us a way to compute
% the unique minimizer of the original problem
%
\begin{rem} \label{rm:remark_after_main_theorem}
Under a few conditions, we can use Theorem \ref{th:agent_agent_coll_max_min} to find
one solution to problem \eqref{eq:coll_prox_op} by solving the simpler problem \eqref{eq:coll_prox_op_single_alpha}
for a special value of $\af$.  In numerical implementations however, the conditions of 
Theorem \ref{th:agent_agent_coll_max_min} are easy to satisfy, and 
the $x^*(\af^*)$ obtained is the unique minimizer of
problem \eqref{eq:coll_prox_op}. We sketch why this is the case in
Appendix \ref{app:remark_after_theorem}.
\end{rem}

%
% simple words describing the theorem
%
In a nutshell, to find one solution to \eqref{eq:coll_prox_op} we simply
find $\af^*$ by maximizing \eqref{eq:expression_for_h}
and then minimize \eqref{eq:coll_prox_op_single_alpha}
using \eqref{eq:x_out_coll_1}-\eqref{eq:x_out_coll_4} with $\af = \af^*$. We can carry both steps efficiently, as shown in Section \ref{sec:num}.
%
% intuition behind the main practical theorem
%
The intuition behind Theorem \ref{th:agent_agent_coll_max_min} is that if we solve the
optimization problem \eqref{eq:coll_prox_op} for the `worst' constraint (the $\af^*$ that gives largest minimum value), 
then the solution also satisfies all other constraints, that is, it holds
for all other $\af \in [0,1]$. We make this precise in the following general lemma that we use to prove Theorem \ref{th:agent_agent_coll_max_min}. We denote by $\partial_i$ the derivative
of a function with respect to the $i^{th}$ variable. The proof of this Lemma is in Appendix
\ref{app:proof_of_main_lemma} and that of Theorem \ref{th:agent_agent_coll_max_min}
is in Appendix \ref{app:proof_of_main_theorem}.
%
% statement of the main general lemma that is required
%
\begin{lem} \label{th:lemma_for_max_min}
Let $\mathcal{A}$ be a convex set in $\mathbb{R}$, $g: \mathbb{R}^d \times \mathcal{A} \rightarrow \mathbb{R}, (x,\af) \mapsto g(x,\af)$, be convex in $\af$ and continuously differentiable in $(x,\af)$ and let $f: \mathbb{R}^d \rightarrow \mathbb{R}, x \mapsto f(x)$, be continuously differentiable and have a unique minimizer. 
For every $\af \in \mathcal{A}$, let $h(\af)$ denote the minimum value of $\min_{x: g(x,\af) \geq 0} f(x)$ 
and if the minimum is attained by some feasible point let this be denoted by $x^*(\af)$.
Under these conditions, if $\af^* \in \arg \max_{\af \in \mathcal{A}} h(\af)$, and if $x^*(\af)$ exists around
a neighborhood of $\af^*$ in $\mathcal{A}$ and is differentiable at $\af^*$, and
if $\partial_1 g(x^*(\af^*),\af^*) \neq 0$, then
$x^*(\af^*)$ minimizes $\min_{x: g(x,\af) \geq 0 \forall \af \in \mathcal{A}} f(x)$.
\end{lem}

%
%*********************************************************************
%
\vspace{-2mm}
\subsection{Other collision operators} \label{sec:other_collision}
\vspace{-0.1cm}
%main idea of this subsection
Using similar ideas to those just described, we now explain how to
efficiently extend to higher dimensions the wall-agent collision operator that
\citet{bento2013message} introduced. In the supplementary video we use
these operators for path planning with obstacles in 3D.

%describe the wall-agent collision minimizer with a single wall
To avoid a collision between agent 1, of radius $r$, and a line between points $y_1,y_2 \in \mathbb{R}^d$, we include the following constraint in the
overall optimization problem: $\| \af x_1(s)   \splus (1 \sminus \af) x_1(s\splus1)  \sminus  (\bt y_1 \splus (1 \sminus \bt ) y_2) \| \geq r$
for all $\af,\bt \in [0,1]$ and all $s\splus1 \in [\eta+1]$. 
This constraint is associated with the proximal operator that receives
$(\un,\un')$ and finds $(\ux,\ox)$ that minimizes
$\urhof \| \ux - \un \|^2 + \orhof \| \ox - \on \|^2$ subject to 
$\| \af \ux \splus (1 \sminus \af) \ox \sminus (\bt y_1  \splus (1 \sminus \bt) y_2 )  \|  \geq r , \text{for all } \af,\bt \in [0,1]$.
Using ideas very similar to those behind Theorem \ref{th:agent_agent_coll_max_min}
and Lemma \ref{th:lemma_for_max_min}, we
solve this problem for dimensions strictly greater than two by maximizing over $\af,\bt \in [0,1]$
the minimum value the single-constraint version of the problem.
In fact, it is easy to generalize 
Lemma \ref{th:lemma_for_max_min} to $\mathcal{A} \subset \mathbb{R}^k$ and 
the single-constraint version of this optimization problem can be obtained from 
\eqref{eq:coll_prox_op_single_alpha} by replacing $\un'$ and $\on'$ with $\bt y_1 \splus (1 \sminus \bt) y_2$,
and letting $\urho', \orho' \rightarrow \infty$. Thus, we use \eqref{eq:expression_for_h} and
\eqref{eq:x_out_coll_1}-\eqref{eq:x_out_coll_4} under this
replacement and limit to generalize the line-agent collision proximal operator of 
\citet{bento2013message}
to dimensions greater than two.
We can also use the same operator to avoid collisions between
agents and a line of thickness $\nu$, by replacing
$r$ with $\nu \splus r$.
%The authors in \cite{bento2013message} seem not to have noticed this possibility.

Unfortunately, we cannot implement a proximal operator to avoid collisions 
between an agent and the convex envelope of an arbitrary set of points $y_1,y_2,...,y_q$
by maximizing over $\af, \bt_1,...,\bt_{q-1} \in [0,1]$ the minimum of the single-constraint
problem obtained from \eqref{eq:coll_prox_op_single_alpha} after replacing
$\un'$ and $\on'$ with $\bt_1 y_1 \splus ... \splus \bt_{q-2} y_{q-2} \splus (1\sminus \bt_1 \sminus ... \sminus \bt_{q\sminus1}) y_{q}$,
and letting $\urho', \orho' \rightarrow \infty$. We can only do so when $d > q$, otherwise
we observe that the condition $\partial_1 g (x^*(\af^*),\af^*)\neq 0$ of
Lemma \ref{th:lemma_for_max_min} does not hold
and $x^*(\af^*)$ is not feasible for the original SIP problem.
In particular, we cannot directly apply our max-min approach
to re-derive the line-agent
collision operator for agents in 2D but only for dimensions $\ge3$.
When $d \leq q$, we believe that a similar but more complicated principle can be
applied to solve the original SIP problem. Our intuition from
a few examples is that this involves considering different portions
of the space $\mathcal{A}$ separately, computing extremal points
instead of maximizing and minimizing and choosing the best feasible solution
among these. We will explore this further in future work.

%we now describe without justification. Let $h(
%\af,\bt_1,...,\bt_{q-1})$ the minimum value of the
%single-constraint optimization problem and $\tilde{h}(
%\af,\bt_1,...,\bt_{q-1})$ the value at the relative minimum of
%the single-constraint optimization problem\footnote{The single-constraint optimization
%problem always has one absolute minimum and one relative minimum. See Appendix
%\ref{app:solving_single_alf_problem}.}. Let $(\af^*,\bt^*_1,...,\bt^*_{q-1})$
%be the least-costly a relative maximum or minimum of either $h$ of $\tilde{h}$ for which the solution
%of the single-constraint problem satisfies 

%then explain that if we want to apply the same results to other
%kind of obstacles we need to be more careful
% in particular the conditions of Lemma ref do not in general hold
% because \| \| = 0, even with a perturbation.

%then explain that the way around it is to replace the max-min principle
%by some sort of extreme extreme principle and check whether, or not,
%there are collisions 

% maybe can say that it is easy, in general to avoid fixed spheres
%

%
%*********************************************************************
%

\subsection{Speeding up computations}
\label{cases}
%\vspace{-.1cm}
% general idea of this section
The computational bottleneck for our collision operators is maximizing \eqref{eq:expression_for_h}.
Here we describe two scenarios, denoted as \emph{trivial} and \emph{easy}, when we
avoid this expensive step to improve performance.

% explain the first cases
First notice that one can readily check whether
$x = n$ is a \emph{trivial} feasible solution. If it is yes, it must be
optimal, because it has $0$ cost, and the operator can
return it as the optimal solution.
This is the case if the segment
from $\Delta \un = \un - \un'$ to $\Delta \on = \on - \on'$ does not intersect
the sphere of radius $r + r'$ centered at zero, which is equivalent 
to $\|\af \Delta \un + (1 - \af )\Delta \on\| \geq r  + r'$ with 
$\af = \max\{1, \min\{0,\af' \} \}$ and
$\af' = \Delta \on^{\dagger} (\Delta \on - \Delta \un) / \|\Delta \on - \Delta \un \|^2$.

%explain the second and third case
The second \emph{easy} case is a shortcut
to directly determine if the maximizer of $\max_{\af \in [0,1]} h(\af)$ is
either $0$ or $1$. We start by noting that empirically
$h$ has at most one
extreme point in $[0,1]$ (the curious reader can convince him/herself of this
by plotting $h(\af)$ for different values of $\Delta \un$ and
$\Delta \on$). This being the case, if $\partial_1 h(0) > 0$ and
$\partial_1 h(1) > 0$ then $\af^* = 1$ and if $\partial_1 h(0) < 0$
and $\partial_1 h(1) < 0$ then $\af^* = 0$. 
Evaluating two derivatives
of $h$ is much easier than maximizing $h$ and can save computation
time. In particular, $\partial_1 h(0) =   C (-(r + r') + \| \Delta \on \|
+ (\Delta \on^{\dagger} (\Delta \un - \Delta \on) /  \|\Delta \on \|))$ and 
$\partial_1 h(1) = C'((r + r') - \| \Delta \un \|
+ (\Delta \un^{\dagger} (\Delta \un - \Delta \on) /  \|\Delta \un \|))$ for 
constants $C,C' > 0$.

%we explain what we do otherwise
If these cases do not hold, we cannot avoid maximizing \eqref{eq:expression_for_h}, a scenario
we denote as \emph{expensive}.
In Section \ref{sec:num} we profile how often each scenario occurs in practice and 
the corresponding gain in speed.

%
%******************************************************
%

\subsection{Local path planning}
\label{local}

% local planning will also improve as a consequence of the new minimizers
The optimization problem \eqref{eq:global_path_opt_problem}
finds beginning-to-end collision-free paths
for all agents simultaneously. This is called global path planning.
It is also possible to solve path planning greedily by solving a sequence of
small optimization problems, i.e. local path planning. Each of these problems plans the
path of all agents for the next $\tau$ seconds
such that, as a group, they get as close as possible to their final desired positions.
This is done, for example, in \citet{fiorini98} and followup work
\cite{alonsomora12icra,alonsomora13icra}. The authors in \citet{bento2013message}
solve these small optimization problems using a special case
of the no-collision operator we study in Section \ref{sec:nocoll} and show this approach
is computationally competitive with the results in \citet{alonsomora13icra}.
Therefore, our results also extend this line of research on local path planning to arbitrary dimensions and 
improve solving-times even further. See Section \ref{sec:num} for details on these improvements in speed.

\vspace{-2mm}

\section{Landmark proximal operator} \label{sec:landmark}

% purpose = free us from having to tell who goes where. as we said before, the frame work introduce dby bento2013 aeasily accounts for the addition of extra stuff. This is the example of something imporatnt to add. the basic idea is that we need to solve three things: (a) representation, (b) assignment and (c) planning. 

%One advantage of planning paths as an optimization
%problem is that we can account for additional concepts in planning
%simply by introducing new terms in the objective function. If the optimization
%problem is solved using a decomposition-coordination proximal algorithm, 
%we only need to compute new proximal operators for the newly introduced terms.

In this section we introduce the concept of \emph{landmarks} that, automatically and jointly, (i) produce reference points in
space-time that, as a group, agents should try to visit, (ii) produce a good assignment between these
reference points and the agents, and
(iii) produce collision-free paths for the agents
that are trying to visit points assigned to them.
%With these new operators we do not to have to specify explicit initial and final points for
%all agents using $f^{\text{pos}}$ in \eqref{eq:global_path_opt_problem}.

% importance = task assignment problems of surveillance or animation (concrete references)
% previous work = OTHERS, task are relatively short, one-shot tasks, and transitions between tasks are irrelevant;  JAVIER, there is no temporal conherence
% katsu work = landmark trajectories computer before hand with points that are already not colliding + assignment is done separately, assuming that robots will follow trajectories perfectly + collisions are not 100% avoided but included using fixed cost + the positive part is that the assinemt is done across trajectories in long sequences
% our work = simulatenous resolution of assignmet and colllision avoidance made with hard constraint and not soft cost. We also consider assignment that is done across long trajectories in long sequences. We solve (1) + (2) and (3) simultaneously
Points (i) to (iii) are essential, for example, to formation control in multi-robot systems
and autonomous surveillance or search \cite{bahceci2003review}, and are also related
to the problem of assigning tasks to robots, if the tasks are seen as groups of points to visit \cite{michael2008distributed}.
Many works focus on
only one of these points or treat
them in isolation. One application where points (i) to (iii) are considered, although separately, 
is the problem of using color-changing robots as pixels in a display \cite{alonso2012image,alonso2012object,flyfire2010}. 
The pixel-robots arrangement is planned frame-by-frame and
does not automatically guarantee that the same image part is represented by the same robots
across frames, creating visual artifacts. Our landmark formalism allows us to penalize these situations.

We introduce landmarks as extra terms in the objective function
\eqref{eq:global_path_opt_problem}; we now explain how to compute
their associated proximal operators.
%
% first we talk about synchronous trajectory landmark
Consider a set of landmark trajectories $\{y_j(s)\}_{j \in [m], s_{\text{init}} \leq s \leq s_{\text{end}}}$ and, to each
trajectory $j$, assign a cost $\tilde{c}_j > 0$, which is the cost of ignoring the entire landmark trajectory. In addition, to each
landmark $y_j(s) \in \mathbb{R}^d$ that is assigned to an agent, assign a penalty $ c_j(s)  > 0$ for deviating from $y_j(s)$.
Landmark trajectories extend the objective function \eqref{eq:global_path_opt_problem}
by adding to it the following term
{\small
\begin{equation} \label{eq:general_syncronized_landmark}
\sum_{j:\sigma_j \neq *} \sum^{s_{\text{end}}}_{s = s_{\text{init}}  } c_j(s) \| x_{\sigma_j}(s) - y_j(s)\|^2
+ \hspace{-0.2cm} \sum_{j:\sigma_j = *} \tilde{c}_j,
\end{equation}
}
where the variable $\sigma_j$ indicates which agent should follow trajectory $j$. If $\sigma_j = *$, that means
trajectory $j$ is unassigned. Each trajectory can be assigned to at most one agent and vice-versa, which it must follow throughout its duration. So we 
have
$\sigma_j \in [p] \cup \{*\}$ as well as the condition that if $\sigma_j = \sigma_{j'}$ then either
$j = j'$ or $\sigma_j = *$. We optimize the overall objective function over $x$ and $\sigma$.
Note that it is not equally important to follow every point
in the trajectory. For example, by setting some $c$'s equal to zero we can effectively
deal with trajectories of different lengths, different beginnings and ends, and even trajectories with holes. 
By setting some of the $c$'s equal to infinity we impose that, if the
trajectory is followed, it must
be followed exactly.
In \eqref{eq:general_syncronized_landmark} we use the Euclidean metric but
other distances can be considered, even non-convex ones, as long
as the resulting proximal operators are easy to compute.
Finally, notice that, a priori, we do \emph{not} need $\{y_j(s)\}$ to describe collision free trajectories.
The other terms in the overall objective function will try to enforce no-collision constraints
and additional dynamic constraints. Of course, if we try to
satisfy an unreasonable set of path specifications, the ADMM or TWA might
not converge.

% now we talk about the proximal operator associated to them
The proximal operator associated to term \eqref{eq:general_syncronized_landmark}
receives as input $\{n_i(s)\}$
and outputs $\{x^*_i(s)\}$ where $i \in [p]$, $s_{\text{init}} \leq s \leq s_{\text{end}}$
and $\{x^*_i(s)\}$ minimizes
{\small
\begin{align} \label{eq:prox_operator_sync_landmarks}
&\min_{x,\sigma} \sum_{j:\sigma_j \neq *} \sum^{s_{\text{end}}}_{s = s_{\text{init}}  } c_j(s) \| x_{\sigma_j}(s) - y_j(s)\|^2+ \hspace{-0.25cm} \sum_{j:\sigma_j = *} \tilde{c}_j \nonumber\\
& + \sum^p_{i = 1} \sum^{s_{\text{end}}}_{s = s_{\text{init}}}
\frac{\rho_i}{2} \| x_i(s)  - n_i(s)\|^2.
\end{align}
}
The variables $\sigma$'s are used only internally in the computation of the proximal
operator because they are not shared with other terms in the overall objective function.
The above proximal operator can be efficiently computed as follows. We first optimize
\eqref{eq:prox_operator_sync_landmarks} over the $x$'s as a function of $\sigma$ and
then we optimize the resulting expression over the $\sigma$'s. If we optimize over the $x$'s
we obtain $\sum_{j} \omega_{j, \sigma_j}$ where, if $\sigma_j = *$,
$\omega_{j, *} = \tilde{c}_j$ and, if $\sigma_j = i \neq *$, then $\omega_{j, i} = 
\min_{x} \sum^{s_{\text{end}}}_{s = s_{\text{init}}} c_j(s) \|x_i(s) - y_j \|^2 + \frac{\rho_i}{2} \|x_i(s) - n_i(s)\|^2 = \sum^{s_{\text{end}}}_{s = s_{\text{init}}} \frac{\rho_i c_j(s)}{2 c_j(s) + \rho_i} \|n_i(s) - y_j(s) \|^2$. The last equality follows from solving a simple quadratic problem.
We can optimize over the $\sigma$'s by solving a linear assignment problem with cost matrix
$\omega$, which can be done, for example, using Hungarian method of \citet{kuhn1955hungarian}, using more advanced methods such as those after
\citet{goldberg1988new}, or using scalable
but sub-optimal algorithms as in \citet{bertsekas1988auction}. Once an optimal
$\sigma^*$ is found, the output of the operator can be computed as follows.
If $i$ is such that $\{j \hspace{-1mm}:\hspace{-1mm} \sigma^*_j  \hspace{-2mm}=\hspace{-2mm}  i \} \hspace{-1mm}=\hspace{-1mm} \emptyset$ then $x^*_i(s) \hspace{-1mm}=\hspace{-1mm} n_i(s)$ for all $s_{\text{init}} \leq s \leq s_{\text{end}}$ and if $i$ is such that $i = \sigma_j$ for some $j \in [m]$ then
$x^*_i(s) = (\rho_i n_i(s) + 2 c_i(s) y_j(s))/ (2 c_i(s) + \rho_i)$ for all $s_{\text{init}} \leq s \leq s_{\text{end}}$.
 
% now we talk about more general trajectories
The term \eqref{eq:general_syncronized_landmark} corresponds to a set
of trajectories between break-points
$s = s_{\text{init}}$ and $s = s_{\text{end}}$ for which the different agents must compete,
that is, each agent can follow at most one trajectory.
We might however want to allow an agent to be assigned to and cover multiple
landmark trajectories.
One immediate way of doing so is by
adding more terms of the form \eqref{eq:general_syncronized_landmark} to the
overall objective function such that the $k^{th}$ term has all its $m^{(k)}$ trajectories
within the interval
$[s^{(k)}_{\text{init}},s^{(k)}_{\text{end}}]$, and different intervals for different $k$'s are
disjoint. However, just doing this does not allow us to impose
a constraint like the following: ``the $j^{th}$ trajectory in the set corresponding to the interval $[s^{(k)}_{\text{init}},s^{(k)}_{\text{end}}]$ must be covered by the same agent as the
the $(j')^{th}$ trajectory in the set corresponding to the interval $[s^{(k+1)}_{\text{init}},s^{(k+1)}_{\text{end}}]$.''
To do so we need to impose the additional constraint that some of the $\sigma^{(k)}$ variables
across different terms of the form \eqref{eq:general_syncronized_landmark} are the same, e.g. 
in the previous example, $\sigma^{(s)}_j = \sigma^{(s+1)}_{j'}$.
Since the variables $\sigma$'s can now be shared across different terms, the
proximal operator \eqref{eq:prox_operator_sync_landmarks} needs to change. 
Now it receives as input a set of values $\{n_i(s)\}_{s,i}$ and $\{n'_j\}_{j}$ and outputs
a set of values $\{x^*_i(s)\}_{i,s}$ and $\{\sigma^*_j\}_j$ that minimize
$ \sum_{j:\sigma_j \neq *} \sum^{s_{\text{end}}}_{s = s_{\text{init}}  } c_j(s) \| x_{\sigma_j}(s) \sminus y_j(s)\|^2 + \hspace{-0.0cm}\sum_{j:\sigma_j = *} \hspace{-0.0cm} \tilde{c}_j + \sum_{i,s} \frac{\rho_i}{2} \| x_i(s)  \sminus n_i(s)\|^2 + \sum^m_{j = 1}  \frac{\rho'_j}{2} \| \sigma_j  \sminus n'_j\|^2$.
%
%\vspace{-0.15cm}
%
%{\small
%\begin{align*}
%&\min_{x,\sigma} \sum_{j:\sigma_j \neq *} \sum^{s_{\text{end}}}_{s = s_{\text{init}}  } c_j(s) \| x_{\sigma_j}(s) \sminus y_j(s)\|^2
%\splus \hspace{-0.2cm}\sum_{j:\sigma_j = *} \hspace{-0.2cm} \tilde{c}_j
%\\
%& \splus \sum_{i,s}
%\frac{\rho_i}{2} \| x_i(s)  \sminus n_i(s)\|^2
 %\splus \sum^m_{j = 1}  \frac{\rho'_j}{2} \| \sigma_j  \sminus n'_j\|^2.
%\end{align*}
%}

In the expression above,
$\{\sigma_j\}_j$ and $\{n'_j\}_j$ are both
vectors of length $p+1$, where the last component encodes for no assignment and
$\sigma_j$ must be binary with only one $1$ entry. For example,
if $p = 5$ and $\sigma_2 = [0,0,1,0,0,0]$ we mean that the second trajectory is assigned to the third agent, or 
if $\sigma_4 = [0,0,0,0,0,1]$ we mean that the fourth trajectory is \emph{not} assigned to any agent. However, $n'$ can have real values and several nonzero components.
%
% how to solve the new minimizer

We also solve the problem above
by first optimizing over $x$ and then over $\sigma$. Optimizing
over $x$ we obtain $\sum_{j} \tilde{\omega}_{j, \sigma_j}$, where 
$\tilde{\omega}_{j, i} = \omega_{j, i} \splus \frac{\rho'_{j}}{2} \|[0,...0,1,0,...,0] \sminus n'_j \|^2 = 
\omega_{j, i} \splus \frac{\rho'_{j}}{2} \|n'_j \|^2 \splus \|1 \sminus n'^{(i)}_j \|^2 \sminus \|n'^{(i)}_j \| ^2 = 
\omega_{j, i} \splus \frac{\rho'_{j}}{2} \|n'_j \|^2 \splus 1 \sminus 2 n'^{(i)}_j$. Given the cost matrix
$\tilde{\omega}$, we find the optimal $\sigma^*$ by solving a linear assignment problem.
Given $\sigma^*$, we compute the optimal $x^*$ using exactly the same expressions
as for \eqref{eq:prox_operator_sync_landmarks}.

% new constraints to force equality among different sections
Finally, to include constraints of the kind $\sigma^{(k)}_j =  \sigma^{(k')}_{j'}$ we add
to the objective a term that takes the value infinity
whenever the constraint is violated and zero otherwise.
This term is associated with a proximal operator 
that receives as input $n'_j = (n'^{(1)}_j,...,n'^{(n)}_j)$ and $n'_{j'}= (n'^{(1)}_{j'},...,n'^{(n)}_{j'})$
and outputs $(\sigma^*_j,\sigma^*_{j'}) \in \arg \min_{\sigma_j = \sigma_{j'}} \frac{\rho_j}{2}\|\sigma_j -  n'_j \|^2 +  \frac{\rho'_{j'}}{2}\|\sigma_{j'} -  n'_{j'} \|^2$. Again $\sigma_{j}$ and $\sigma_{j'}$ are binary vectors of length $p+1$ with
exactly one non-zero entry. The solution has the form 
$\sigma^*_j =  \sigma^*_{j'} = [0,0,...,0,1,0,...0]$ where the $1$ is in position $i^* = \arg \max_{i \in [p]} \rho_j n'^{(i)}_j +\rho'_{j'} n'^{(i)}_{j'}$.

%\vspace{-2mm}
\section{Numerical experiments} \label{sec:num}

We gathered all results with a Java implementation of the ADMM and the TWA as described in \citeauthor{bento2013message} (\citeyear{bento2013message}; see Appendix \ref{app:TWA_implementation})
using JDK7 and Ubuntu v12.04 run on a desktop machine with 2.4GHz cores.

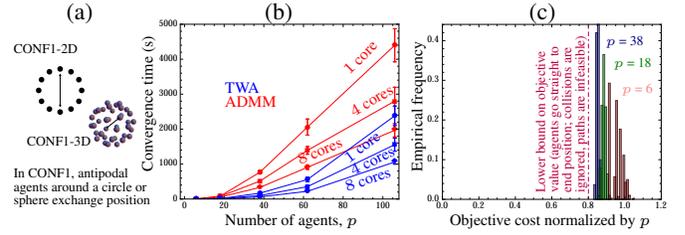
\begin{figure}[h!]
\begin{center}
\includegraphics[trim = 2.3mm -90mm 25mm -3mm, clip,width=0.16\columnwidth]{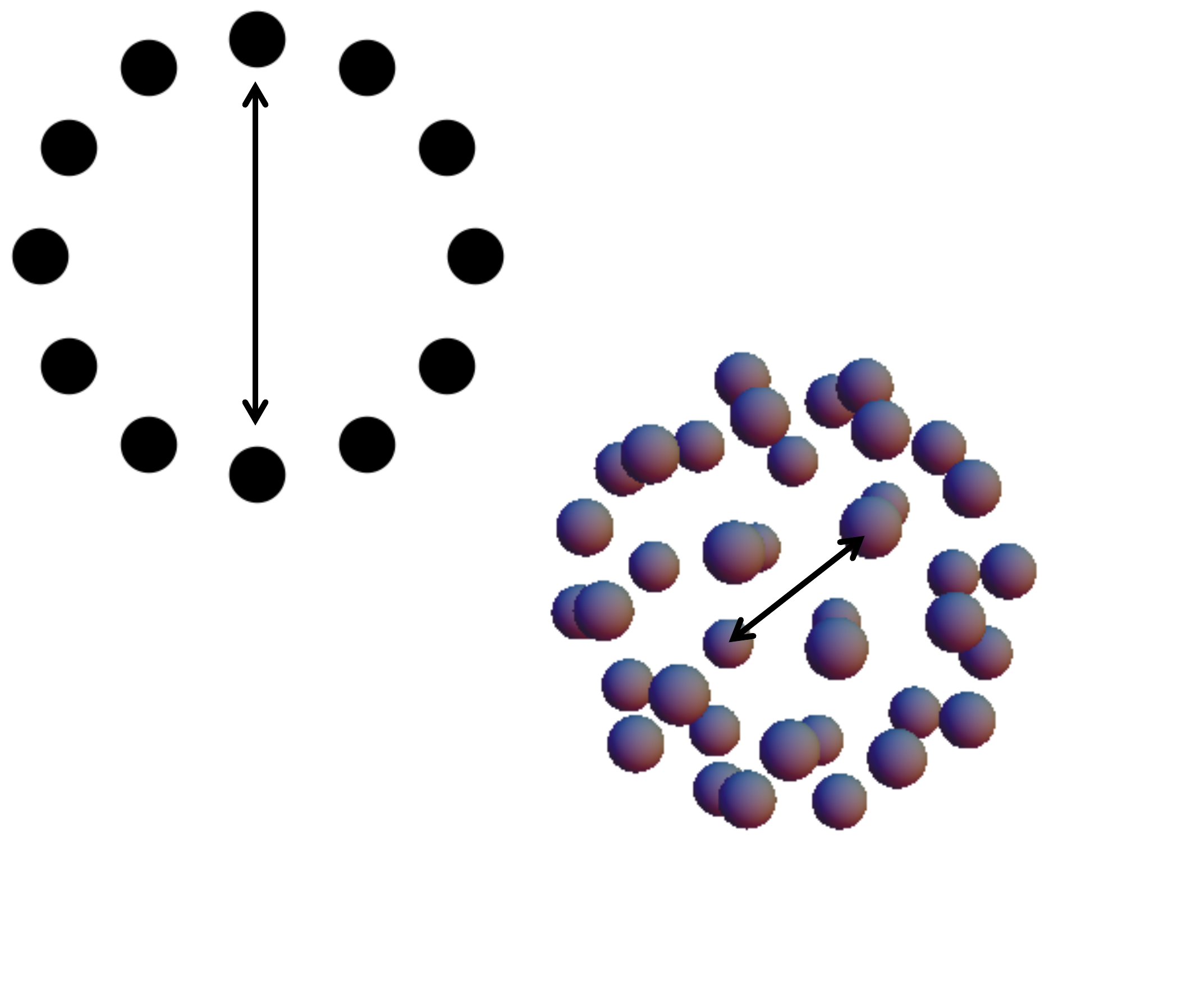}\;\;
\includegraphics[width=0.38\columnwidth]{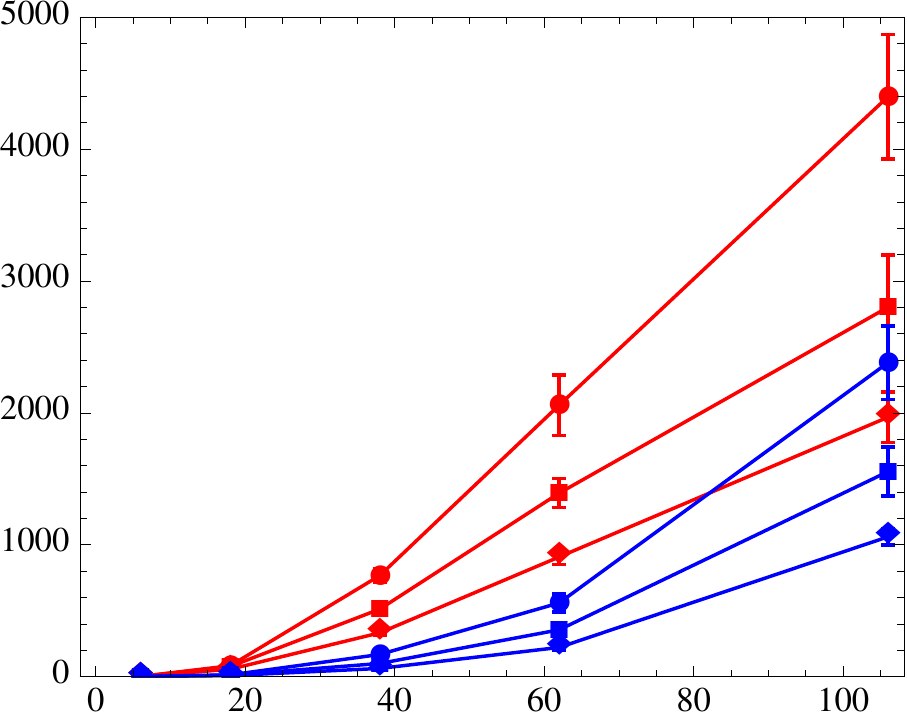} \;\;
\includegraphics[width=0.38\columnwidth]{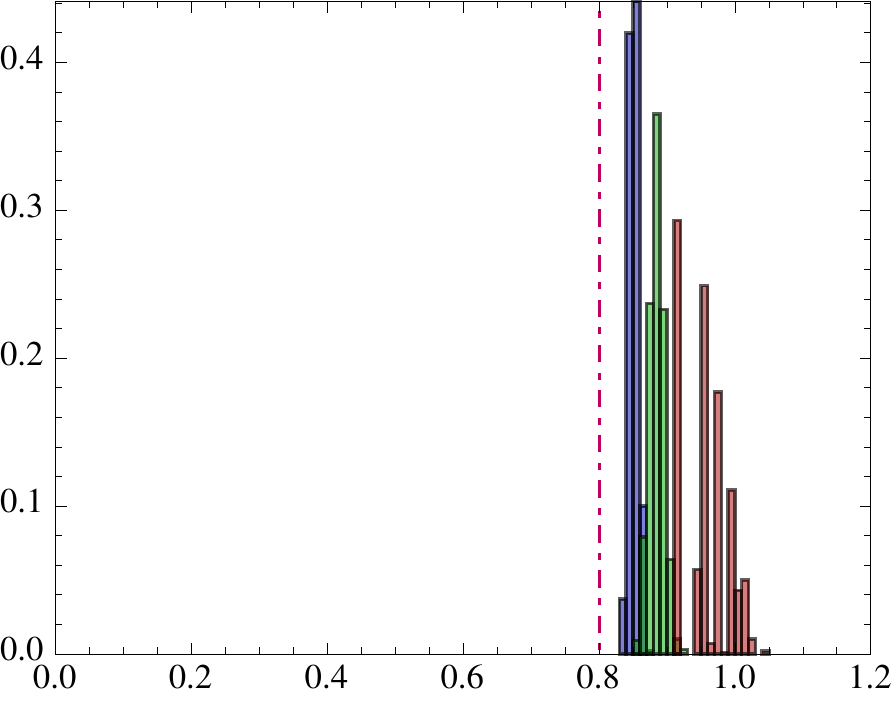}
\put(-230,73){\mbox{\rotatebox{0}{  \small (a) }}}
\put(-155,73){\mbox{\rotatebox{0}{  \small (b) }}}
\put(-55,73){\mbox{\rotatebox{0}{  \small (c) }}}
\put(-50,5){\mbox{\rotatebox{90}{  \fontsize{1.8mm}{1em}\selectfont \color{purple} Lower bound on objective}}}
\put(-45,5){\mbox{\rotatebox{90}{  \fontsize{1.8mm}{1em}\selectfont  \color{purple} value (agents go straight to}}}
\put(-40,5){\mbox{\rotatebox{90}{  \fontsize{1.8mm}{1em}\selectfont \color{purple} end position; collisions are }}}
\put(-35,5){\mbox{\rotatebox{90}{  \fontsize{1.8mm}{1em}\selectfont  \color{purple} ignored, paths are infeasible) }}}
\put(-20,45){\mbox{\rotatebox{0}{  \fontsize{1.8mm}{1em}\selectfont \color{red!50!} $p$ = 6 }}}
\put(-23,55){\mbox{\rotatebox{0}{  \fontsize{1.8mm}{1em}\selectfont \color{green!50!black} $p$ = 18 }}}
\put(-26,63){\mbox{\rotatebox{0}{  \fontsize{1.8mm}{1em}\selectfont \color{blue!50!black} $p$ = 38 }}}
\put(-85,-5){\mbox{\rotatebox{0}{  \tiny  Objective cost normalized by $p$}}}
\put(-97,13){\mbox{\rotatebox{90}{  \tiny Empirical frequency }}}
\put(-170,-5){\mbox{\rotatebox{0}{  \tiny Number of agents, $p$}}}
\put(-199,13){\mbox{\rotatebox{90}{  \tiny Convergence time (s)}}}
\put(-170,40){\mbox{\rotatebox{0}{  \color{red} \tiny ADMM}}}
\put(-170,45){\mbox{\rotatebox{0}{  \color{blue} \tiny TWA}}}
\put(-125,50){\mbox{\rotatebox{45}{  \color{red} \tiny $1$ core}}}
\put(-123,35){\mbox{\rotatebox{30}{  \color{red} \tiny $4$ cores}}}
\put(-143,17){\mbox{\rotatebox{23}{  \color{red} \tiny $8$ cores}}}
\put(-125,7){\mbox{\rotatebox{19}{  \color{blue} \tiny $8$ cores}}}
\put(-123,13){\mbox{\rotatebox{25}{  \color{blue} \tiny $4$ cores}}}
\put(-125,17){\mbox{\rotatebox{37}{  \color{blue} \tiny $1$ core}}}
\put(-250,60){\mbox{\rotatebox{0}{  \fontsize{1.8mm}{1em}\selectfont CONF1-2D}}}
\put(-245,25){\mbox{\rotatebox{0}{  \fontsize{1.8mm}{1em}\selectfont CONF1-3D}}}
\put(-250,13){\mbox{\rotatebox{0}{  \fontsize{1.8mm}{1em}\selectfont In CONF1, antipodal}}}
\put(-250,8){\mbox{\rotatebox{0}{  \fontsize{1.8mm}{1em}\selectfont agents around a circle or}}}
\put(-250,3){\mbox{\rotatebox{0}{  \fontsize{1.8mm}{1em}\selectfont sphere exchange position}}}

%
%\vspace{-0.15cm}
%
\caption{(a) CONF1-2D \& 3D; (b) Convergence time for CONF1-3D for a varying number of cores and agents; (c) Empirical distribution 
of the objective over $1000$ random initializations of TWA
for CONF1-3D.}
\label{fig:numerical-scale3d}
\end{center}
\end{figure}

%speed test for 2D of old vs new minimizer
We first compare the speed of the implementation of the collision operator
as described in this paper, which we shall refer to as ``NEW,'' with
the implementation described in \citet{bento2013message}, which we denote ``OLD.''
We run the TWA using OLD on the 2D scenario
called ``CONF1'' in \citet{bento2013message} with $p=8$ agents of radius $r=0.918$, equally spaced around a circle of radius $R=3$, each required to exchange position with the corresponding antipodal agent (cf. Fig. \ref{fig:numerical-scale3d}-(a)).
While running the TWA using OLD, we record the trace of all $n$ variables input into the OLD operators.
We compare the execution speed of OLD and NEW on this trace of inputs,
after segmenting the $n$ variables into \emph{trivial}, \emph{easy}, or \emph{expensive} according to \S\ref{cases}.
For global planning, the distribution of \emph{trivial}, \emph{easy}, \emph{expensive} inputs is $\{0.814, 0.001, 0.185\}$. 
Although the \emph{expensive} inputs are infrequent, the total wall-clock time that
NEW takes to process them is $76$ msec compared to $54$ msec to process all \emph{trivial} and \emph{easy} inputs. 
By comparison, OLD takes a total time of
$551$ msec on the \emph{expensive} inputs and so our new implementation yields 
an average speedup of $7.25\times$ on the inputs that are most expensive to process.
Similarly, we collect the trace of the $n$ variables input into the collision operator when using the local planning
method described in \citet{bento2013message}
on this same scenario.
We observe
a distribution of the \emph{trivial}, \emph{easy}, \emph{expensive} inputs equal to
$\{0.597,0.121,0.282\}$, we get a total time spent in the \emph{easy} and \emph{trivial} cases of $340$ msec for NEW 
and a total time spent in the \emph{expensive} cases of $2802$ msec for NEW and $24157$ msec for OLD. This is an average speedup of $8.62\times$ on the \emph{expensive} inputs.
For other scenarios, we observe similar speedup on the \emph{expensive} inputs, although 
scenarios easier than CONF1 normally have fewer \emph{expensive} inputs. E.g.,
if the initial and final positions are chosen at random instead of according to CONF1, this distribution
is $\{0.968,0,0.032\}$.

%behavior of new minimizer of higher dimensions
%Although our main focus is not to study the performance of the framework proposed 
%in \cite{bento2013message}, rather we wish to provide a new set of building blocks for it,
Figure \ref{fig:numerical-scale3d}-(b) shows the convergence time for
instances of CONF1 in 3D (see Fig. \ref{fig:numerical-scale3d}-(a)) using NEW for a different
number of agents using both the ADMM and the TWA. We recall that OLD cannot be applied to agents in 3D.
Our results are similar to those
in \citet{bento2013message} for 2D:  (i) convergence time seems to grow
polynomially with $p$; (ii) the TWA is faster than the ADMM; and, (iii) the proximal operators lend
themselves to parallelism, and thus added cores decrease
time (we see $\sim2\times$ with 8 cores).
In Figure \ref{fig:numerical-scale3d}-(c) we show that the paths found when the TWA solves CONF1
in 3D over $1000$ random initializations are not very different and seem to be good (in terms of objective value).
\vspace{-0.15cm}

\vspace{-3mm}
%some comment about the landmarks
In the supplementary movie we demonstrate the use of
the landmark operators.
First we show the use of these operators on six toy problems involving
two agents and four landmark trajectories where
we can use intuition
to determine if the solutions found are good or bad.
We solve these six scenarios using the ADMM with
$100$ different random initializations to avoid local minima and
reliably find very good solutions. With 1 core it always takes less than $3$ seconds to converge and typically less than $1$ second.
We also solve
a more complex problem involving $10$ agents and about $100$ landmarks
whose solution is a `movie' where the different robots act as pixels.
With our landmark operators we do not have to
pre-assign the robots to the pixels in each frame.

\vspace{-2mm}
\section{Conclusion} \label{sec:concl}

%summary of our contributions
We introduced two novel proximal operators
that allow the use of proximal algorithms
to plan paths for agents in 3D, 4D, etc. and also to 
automatically assign waypoints to agents. The growing
interest in coordinating large swarms of quadcopters
in formation, for example, illustrates the importance
of both extensions.
For agents in 2D, our collision operator is
substantially faster than its predecessor.
In particular, it leads to an implementation
of the velocity-obstacle local planning method that is faster than 
its implementation in both \citet{alonsomora13icra} and \citet{bento2013message}.
The impact of our work goes beyond path planning. We are currently working on two other projects that use our results. One is related to visual tracking of multiple non-colliding large objects and the other is related to the optimal design of layouts, such as for electronic circuits. In the first, the speed of the new no-collision operator is crucial to achieve real-time performance and in the second we apply Lemma \ref{th:lemma_for_max_min} to derive no-collision operators for non-circular objects.

%future work
The proximal algorithms used can get stuck in local minima, although empirically
we find good solutions even for hard instances with very
few or no random re-initializations. Future
work might explore improving robustness,
possibly by adding a simple method to start the TWA or the ADMM
from a `good' initial point. Finally, it would be valuable
to implement wall-agent collision proximal operators that are more general
than what we describe in Section \ref{sec:other_collision},
perhaps by exploring other methods to solve SIP problems.

\newpage
\bibliographystyle{aaai}
\bibliography{multi_dimension_robot_planning_with_landmarks}

\newpage

%
%*********************************************************************
%

\section*{\Large Appendix for ``Proximal Operators for Multi-Agent Path Planning''}

%
%*********************************************************************
%

\appendix

%
%******************************
%
\section{A comment on the impact of the assumption of piece-wise linear paths in practice} \label{app:path_smoothness}

A direct application of the approach of \citet{bento2013message} can
result in very large accelerations at the break-points.  
It is however not hard to overcome this apparent limitation in practical applications. We now explain one way of doing it. First notice that by increasing the number of break-points we can obtain trajectories arbitrarily close to smooth trajectories with finite acceleration everywhere. Since in practice it is not efficient to work with a very large number of break-points, we can keep the number of break-points at a reasonable level and increase the effective radius of the robots. This would allow us to fit a polynomial through the break-points and obtain smooth trajectories that are never distant from the piece-wise linear paths by more than the difference between the true robots radii and their effective radii. Using this approach, we would obtain a set of non-colliding finite-acceleration trajectories. We can also impose specific maximum-acceleration constrains if, at the same time, we restrict the maximum permitted change of velocity at break-points using additional proximal operators.

\section{An illustration of the two kinds of blocks used by the ADMM/TWA}
\label{app:ADMM_TWA_blocks}

As explained in Section \ref{sec:backG}, the ADMM/TWA is an iterative
scheme that alternates between (i) producing different estimates
of the optimal value of the variables each function in the objective
depends on and (ii) producing consensus values from the different
estimates that pertain the same variable. The blocks that produce
the estimates we call proximal operators and the blocks that
produce consensus values we call consensus operators.
The proximal operator blocks only receive messages from the consensus
blocks (and send estimates back to them) and the consensus blocks
only receive estimates from the proximal operators (and send consensus
messages back to the proximal operators).
Hence, the ADMM iteration scheme can be interpreted as
messages passing back and forth along the edges of a bipartite graph.

\begin{figure}[h!]
\begin{center}
\includegraphics[trim = 0mm 0mm 0mm 0mm, clip, width=7cm]{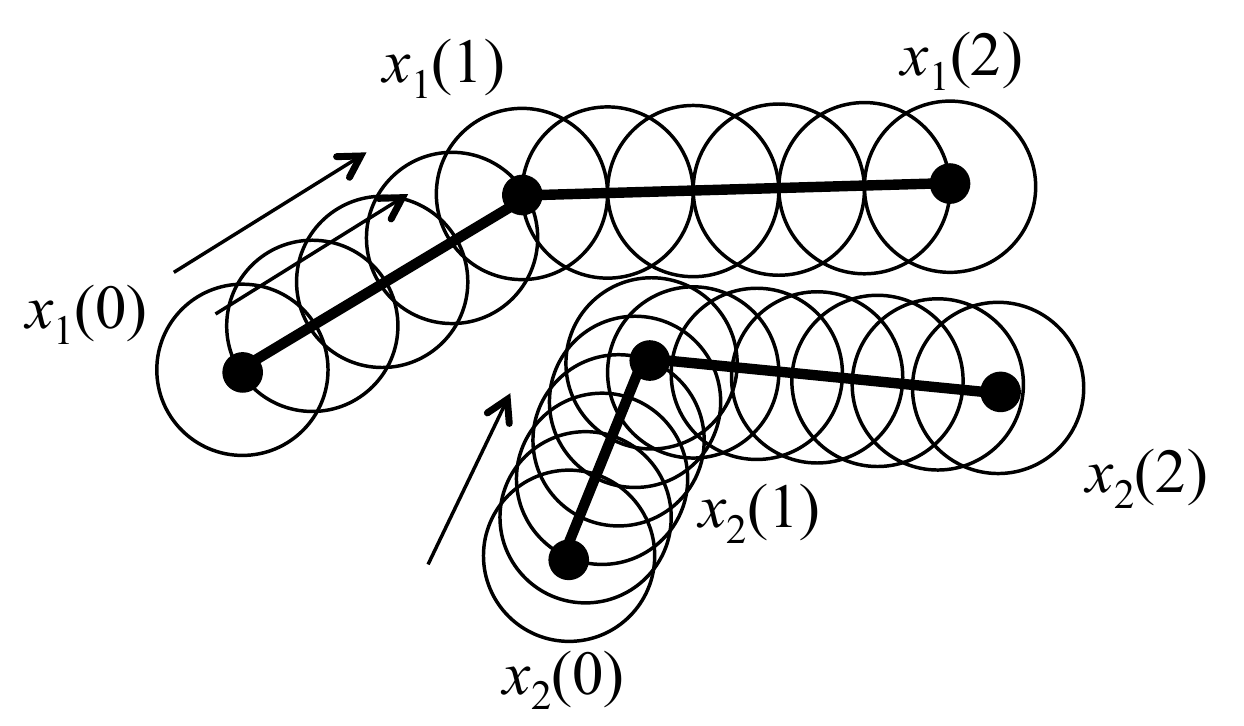}\\ \vspace{0.5cm}
\includegraphics[trim = 0mm 0mm 0mm 0mm, clip, width=7cm]{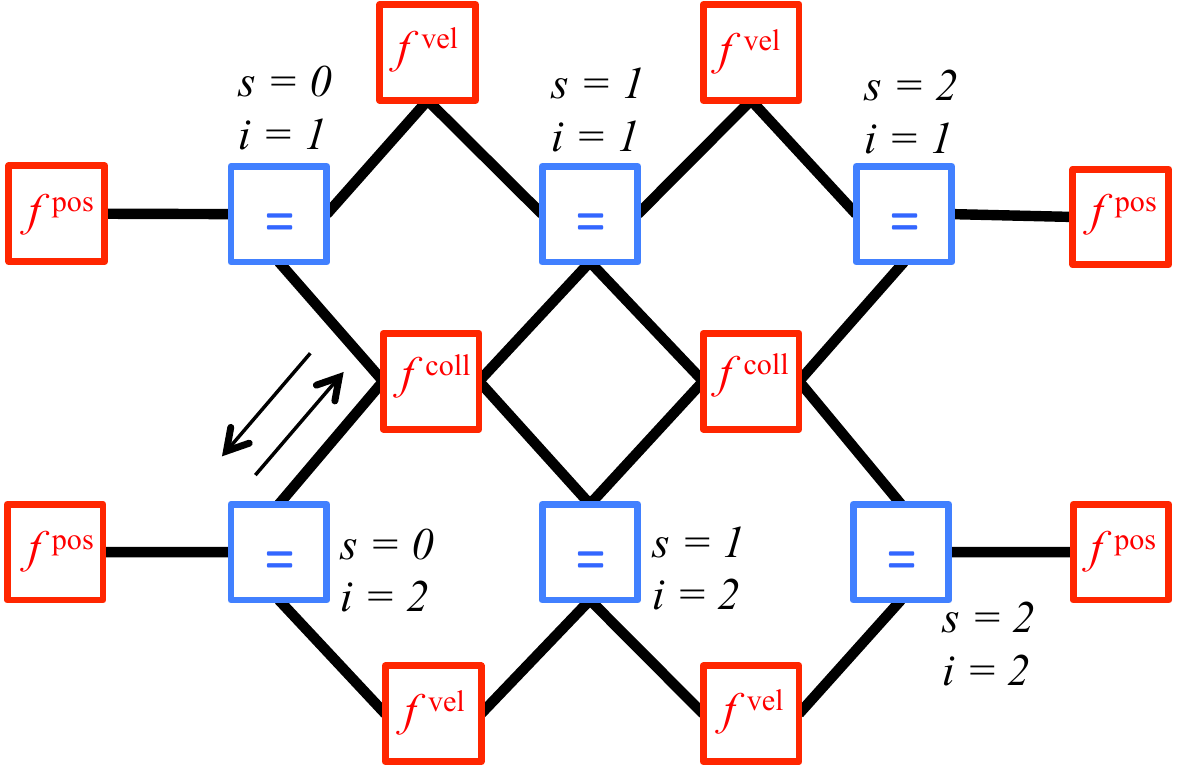}\\ \vspace{0.5cm}
\includegraphics[trim = 0mm 0mm 0mm 0mm, clip, width=7cm]{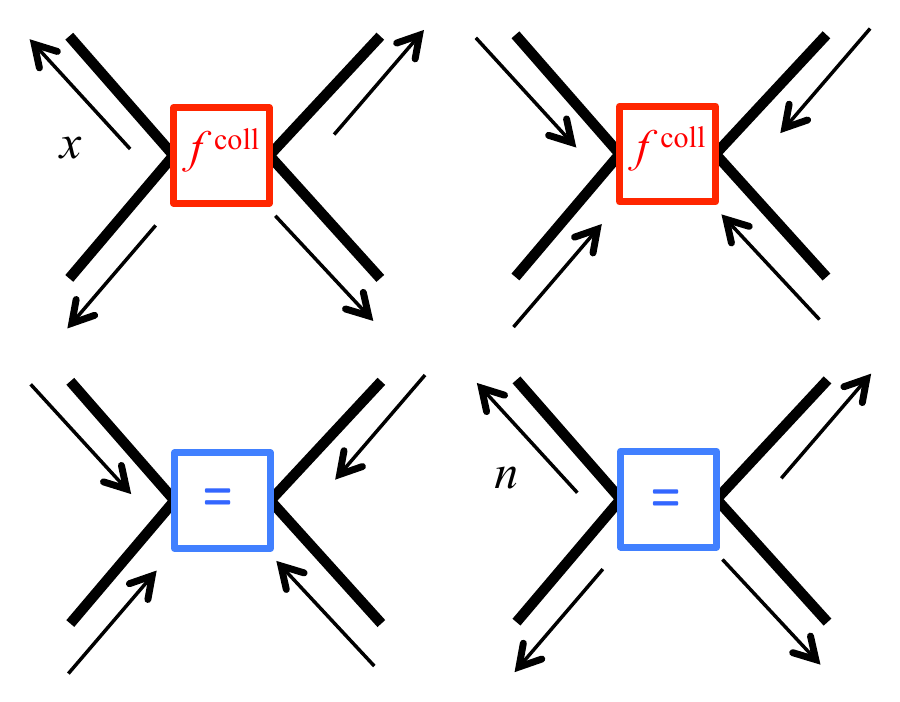}
%
%\put(-200,200){(a)}
%\put(-220,200){(b)}
%\put(-220,140){(c)}
%
\caption{(Top) Variables in the problem; (Center) Graph of connections between  proximal operators and consensus nodes; (Bottom) Proximal operators' input and output values.}
\label{fig:simple_instance}
\end{center}
\end{figure}

We now illustrate this. Imagine that we want to compute non-colliding paths
for two agents and that trajectories are parametrized by three break-points.
In this case the optimization problem \eqref{eq:global_path_opt_problem}
has six variables, namely, $x_1(0),x_1(1),x_1(2)$ for agent $1$ and 
$x_2(0),x_2(1),x_2(2)$ for agent $2$. See Figure \ref{fig:simple_instance}-(top).

The ADMM has one consensus operator associated with the
position of each agent at each break-point (blue blocks with `=' sign on it)
and has one proximal operator associated with each function in the
objective (red blocks with function names on it).
In our small example, we have two no-collision operators.
One ensures there are no-collisions between the segment
connecting $x_1(0)$ and $x_1(1)$ and the segment connecting
$x_2(0)$ and $x_2(1)$. The other acts similarly on 
$x_1(1)$, $x_1(2)$, $x_2(1)$ and $x_2(2)$. We also have
four velocity operators that penalize trajectories in which
the path segments have large velocities. Finally, we have
four position proximal operators that enforce agent $1$ and
$2$ to start and finish their paths at specified locations.
See Figure \ref{fig:simple_instance}-(center).

The crucial part of implementing the ADMM or the TWA is
the construction of the proximal operators. The proximal
operators receive consensus messages $n$ from the
consensus nodes and produce estimates for the values of
the variables of the function associated to them. For example,
at each iteration, the left-most no-collision proximal operator in 
Figure \ref{fig:simple_instance}-(center) receives messages $n$
from the consensus nodes associated to the variables
$x_1(0)$, $x_1(1)$, $x_2(0)$ and $x_2(1)$ and produces
new estimates for their optimal value. The center-top consensus
operator receives fours estimates for $x_1(1)$, from two no-collision
proximal operators and two velocity proximal operators, and
produces a single estimate its optimal value. See
Figure \ref{fig:simple_instance}-(bottom).

%
%*********************************************************************
% in this section we explain the remark that is done after the statement of Theorem 1
\section{A comment on our implementation of the TWA}
\label{app:TWA_implementation}

Implementing the TWA requires computing proximal
operators and specifying, at every iteration, what \citet{nate13} calls the
\emph{outgoing weights}, $\overrightarrow{\rho}$, of each proximal
operator. In the TWA there are also \emph{incoming weights},
$\overleftarrow{\rho}$, which correspond to the $\rho$'s that appear in the
definition of all our proximal operators, but their update scheme
at every iteration is fixed \cite[Section 4.1]{nate13}.

For the collision operator and landmark operator we introduce
in this paper, we compute the outgoing weights using the same
principle as in \cite{bento2013message}. Namely, if an operator maps a set
of input variables $(n_1,n_2,...,n_k)$ to a set of output variables
$(x_1,x_2,...,x_k)$ then, if $n_i \neq x_i$ we set $\overrightarrow{\rho}_i = 1$
otherwise, when $n_i = x_i$, we set $\overrightarrow{\rho}_i = 0$. In other words,
if a variable is unchanged by the operators, the outgoing weight associated with it
should be zero, otherwise it is $1$.

%
%*********************************************************************
% in this section we explain the remark that is done after the statement of Theorem 1
\section{More details about Remark \ref{rm:remark_after_main_theorem} }
\label{app:remark_after_theorem}

% enumerate a few observations
The next three points sketch why Remark \ref{rm:remark_after_main_theorem} is true.

%minimizer for original problem exists
First, if the $\rho$'s are all positive, the objective function of problem \eqref{eq:coll_prox_op}
is strictly convex and we can add the constraint
$\|(\ux,\ux',\ox,\ox')\| \leq M$, for $M$ large enough, without changing
its minimum value. The new extended set of constraints amounts to the intersection
of closed sets with a compact set and by the continuity of the objective function
and the extreme value theorem it follows that the problem has a minimizer.

% in general there is only one solution to the original problem
Second, problem \eqref{eq:coll_prox_op} can be interpreted as
resolving the following conflict.
``Agent 1 wants to move from $\un$ at time $0$ to $\on$ at time $1$ and agent
2 wants to move from $\un'$ at time $0$ to $\on'$ at time $1$,
however, if both move in a straight-line, they collide.
How can we minimally perturb their initial and final reference positions so
they they avoid collision?'' If the vectors $(\un,0)$, $(\un',0)$, $(\on,1)$,$(\on',1)$
lie in the same three-dimensional plane there can be ambiguity on how
to minimally perturb the agents' initial and final positions: agent 1's  reference
positions can either move `up' and agent 2's `down'
or agent 1's reference positions `down' and  agent 2's `up'
(`up' and `down' relative to the plane defined by the vectors $(\un,0)$, $(\un',0)$, $(\on,1)$,$(\on',1)$).
In numerical implementations however, it almost never happens that 
$(\un,0)$, $(\un',0)$, $(\on,1)$,$(\on',1)$ lie in the same plane and, in fact, 
this can be avoided by adding a very small amount of random noise to the $n$'s before
solving problem \eqref{eq:coll_prox_op}.

% observation about the existen0ce of a solution to the second problem
Third, one can show from the continuity the objective function of problem \eqref{eq:coll_prox_op}
, the continuous-differentiability of $g(x,\af) = \| \af (\ux - \ux') + (1 - \af) (\ox - \ox') \|^2 - (r+r')^2$
and the fact that the level sets of $g(x,\af)$ as a function of $x$ never have `flat' sections
that $h(\af)$ is a continuous function. Since $[0,1]$ is compact, it follows that there always exists an $\af^*$.
Also, for the purpose of a numerical implementation, we can consider
$\| \af^* (\un - \un') + (1 - \af^*) (\on - \on') \| \neq 0$.
In fact, this can be avoided by adding a very small amount of
random noise to the $n$'s before solving problem \eqref{eq:coll_prox_op_single_alpha}.
Therefore, for practical purposes,
we can consider that for each $\af^*$ there exists a unique minimizer to
problem $\eqref{eq:coll_prox_op_single_alpha}$.
Finally, a careful inspection of \eqref{eq:expression_for_h} shows that
if there exists $\af$ such that $h(\af) > 0$ then $\af^*$ is unique and
if $h(\af) = 0$ for all $\af$ then
\eqref{eq:x_out_coll_1}-\eqref{eq:x_out_coll_4} always give $x = n$.
In short, for the purpose of a numerical implementation,
we always find a unique $x^*(\af^*)$ and because, in practice, as argued in the two points above,
problem \eqref{eq:coll_prox_op} can be considered to have only one solution, it
follows that this unique $x^*(\af^*)$ is 
the unique minimizer of problem \eqref{eq:coll_prox_op}.

%
%*********************************************************************
% in this section we prove our main lemma
\section{Proof of Lemma \ref{th:lemma_for_max_min}} \label{app:proof_of_main_lemma}

We need to consider two separate cases.

%
% we prove for the first case, the easier case
%
In the first case we assume that $x^*(\af^*)$ is such that
$g(x^*(\af^*),\af^*) > 0$. This implies that $x^*(\af^*)$ is a minimizer of $\min_x f(x)$, which implies that $\min_{x: g(x,\af^*) \geq 0} f(x) = \min_x f(x)$, which implies that $\min_{x: g(x,\af) \geq 0} f(x) \geq \min_x f(x) = \min_{x: g(x,\af^*) \geq 0} f(x) = \max_{\af' \in \mathcal{A}} \min_{x: g(x,\af') \geq 0} f(x) \geq \min_{x: g(x,\af) \geq 0} f(x)$.
In other words, $\min_{x: g(x,\af) \geq 0} f(x) = f(x^*(\af^*))$ for all $\af \in \mathcal{A}$. Since $f(x)$ has a unique minimizer we have that
%$x^*(\af)= x^*(\af^*)$ for all $\af \in \mathcal{A}$, which implies that
$x^*(\af^*)$ must be feasible for the problem $\min_{x: g(x,\af) \geq 0} f(x)$, which implies that $g(x^*(\af^*),\af) \geq 0$ for all $\af \in \mathcal{A}$. In other words, $x^*(\af^*)$ is feasible point of $\min_{x: g(x,\af) \geq 0 \forall \af \in \mathcal{A}} f(x)$ and attains the smallest possible objective value, hence it minimizes it.

%
% we prove for the second case, first we give a high-level view
%
In the second case we assume that $g(x^*(\af^*),\af^*) = 0$. 
To finish the proof it suffices to show 
that $v \partial_2 g (x^*(\af^*),\af^*) \geq 0$ for all $v \in \mathbb{R}$ such that $\af^* + v \in \mathcal{A}$.
To see this, we first notice that, if this is the case, then, for all $\af \in \mathcal{A}$,
we have by convexity that $g (x^*(\af^*),\af) \geq g (x^*(\af^*),\af^*) + (\af - \af^*) \partial_2 g (x^*(\af^*),\af^*) = (\af - \af^*) \partial_2 g (x^*(\af^*),\af^*) \geq 0$, which implies that $x^*(\af^*)$ is a feasible point for the problem
$\min_{x: g(x,\af) \geq 0, \forall \af \in \mathcal{A}} f(x)$. Secondly, we notice that $f(x^*(\af^*)) \geq \min_{x: g(x,\af) \geq 0, \forall \af \in \mathcal{A}} f(x) \geq \max_{\af \in \mathcal{A}} \min_{x: g(x,\af) \geq 0} f(x) = f(x^*(\af^*))$. This implies that $x^*(\af^*)$ minimizes $\min_{x: g(x,\af) \geq 0, \forall \af \in \mathcal{A}} f(x)$.

%
% now we give the details that were missing above in the second case
%
We now show that $v \partial_2 g (x^*(\af^*),\af^*) \geq 0$
for all $v \in \mathbb{R}$ such that $\af^* + v \in \mathcal{A}$.

First we notice that since $f$ is differentiable and $x^*(\af)$ exists and is differentiable
at $x^*(\af^*)$, then, by the chain rule, $h(\af)$ is differentiable at $\af^*$.
In addition, we notice that if $\af^*$ maximizes  $h$ then, if $v \in \mathbb{R}$ is such that $\af^* + v \in \mathcal{A}$, 
the directional derivative of $h$ in the direction of $v$
evaluated at $\af^*$ must be non-positive. In other words,
$v \partial_1 h(\af^*) = v \partial_1 f(x^*(\af^*))^{\dagger} \partial_1 x^*(\af^*) \leq 0$.
Second, we notice that in a small neighborhood around $\af^*$, 
$x^*(\af)$ exists and is continuous (because $x^*(\af^*)$ is differentiable) which,
by the continuous-differentiability
of $g$ and the fact that $\partial_1 g(x^*(\af^*),\af^*) \neq 0$,
implies that $\partial_1 g(x^*(\af),\af) \neq 0$ in this neighborhood.
Therefore, in a small neighborhood around $\af^*$,
the problem $\min_{x: g(x,\af) \geq 0} f(x)$ has a single inequality constraint
and $\partial_1 g(x^*(\af),\af) \neq 0$
which implies that $x^*(\af)$, which we are assuming exists in this neighborhood,
is a feasible \emph{regular point} and satisfies the first-order necessary optimality conditions \cite{bertsekas1999nonlinear}\footnote{Also known as also known as Karush-Kuhn-Tucker (KKT) conditions.} 
\begin{align}
\partial_1 f (x^*(\af)) + \lambda \partial_1 g(x^*(\af),\af) = 0,\label{eq:main_kkt_condition_general_problem}\\
\lambda g (x^*(\af),\af) = 0,\\
g (x^*(\af),\af) \geq 0, \label{eq:inequ_const_general_problem}\\
\lambda \leq 0.
\end{align}
Now, we take the directional derivative of
\eqref{eq:inequ_const_general_problem} with respect to $\af$ in the direction $v$ evaluated at $(x^*(\af^*),\af^*)$ and
obtain
\begin{equation} \label{eq:perturbing_the_constrain}
v \partial_1 g(x^*(\af^*),\af^*)^{\dagger} \partial_1 x^*(\af^*) + v \partial_2 g(x^*(\af^*),\af^*) \geq 0.
\end{equation}
At the same time, by computing the inner product between \eqref{eq:main_kkt_condition_general_problem} and $\partial_1 x^*$ evaluated at $\af^*$ and multiplying by $v$ we obtain $v \partial_1 f(x^*(\af^*))^\dagger \partial_1 x^*(\af^*) + \lambda v \partial_1 g(x^*(\af^*),\af^*)^\dagger \partial_1 x^*(\af^*) = 0$. But we have already proved that $v \partial_1 f(x^*(\af^*))^{\dagger} \partial_1 x^*(\af^*) \leq 0$ therefore
$\lambda v \partial_1 g(x^*(\af^*),\af^*)^\dagger \partial_1 x^*(\af^*) \geq 0$. Now recall that we are assuming $g(x^*(\af^*),\af^*) = 0$ therefore, $\lambda < 0$. It thus follows that $v \partial_1 g(x^*(\af^*),\af^*)^\dagger \partial_1 x^*(\af^*) \leq 0$ and from \eqref{eq:perturbing_the_constrain} we conclude that 
$v \partial_2 g(x^*(\af^*),\af^*) \geq 0$.

%
%*********************************************************************
% in this section we prove theorem 1 using lemma 1
\section{Proof of Theorem \ref{th:agent_agent_coll_max_min}} \label{app:proof_of_main_theorem}

% first observe some modification
We first observe that the solutions to problem \eqref{eq:coll_prox_op} and \eqref{eq:coll_prox_op_single_alpha}
remain the same if we replace $\|\af (\ux - \ux') + (1 - \af) (\ox - \ox')\| \geq r +r' $ by
$\|\af (\ux - \ux') + (1 - \af) (\ox - \ox')\|^2  - (r +r')^2 \geq 0$. We prove the theorem with this replacement in
mind.

% general structure of the theorem
We first prove the theorem assuming that we have proved that
that expressions \eqref{eq:expression_for_h}
to \eqref{eq:x_out_coll_4} hold when $\| \af (\un - \un') + (1 - \af) (\on - \on') \| \neq 0$.
This is then proved last.

% need to consider two separate cases
To prove the theorem we consider two separate cases.

% first case
In the first case we consider $r + r' \leq \min_{\af \in [0,1]} \|\af (\un - \un') + (1 - \af) (\on - \on')\|$.
This implies that the minimum of problem \eqref{eq:coll_prox_op} is $0$ and, because
the $\rho$'s are positive, there is a unique minimizer which equals $(\un,\un',\on,\on')$.
In this case we also have $h(\af) = 0$ for all $\af$, which implies that the solution of problem
\eqref{eq:coll_prox_op_single_alpha} is $x^*(\af) = n$ for all $\af$.
Therefore, for any optimal $\af^*$ for which $\| \af^* (\un - \un') + (1 - \af^*) (\on - \on') \| \neq 0$, we have
that the minimizer of problem \eqref{eq:coll_prox_op_single_alpha} is equal to the 
unique solution of problem \eqref{eq:coll_prox_op}. Hence the theorem is true in this case.

%second case
Now we assume that $r + r' > \min_{\af \in [0,1]} \|\af (\un - \un') + (1 - \af) (\on - \on')\|$.
This implies that there exists an $\af$ for which the right-hand-side of \eqref{eq:expression_for_h} is positive
and for which $\| \af (\un - \un') + (1 - \af) (\on - \on') \| \neq 0$. This implies that, there exists
an $\af$ for which $h(\af) > 0$. Therefore, for any optimal $\af^*$ it must the case that $h(\af^*) > 0$.
If $\| \af^* (\un - \un') + (1 - \af^*) (\on - \on') \| \neq 0$ then \eqref{eq:expression_for_h} holds around a neighborhood
of $\af^*$ and in this neighborhood $h(\af) > 0$. Therefore, in this neighborhood $x^*(\af)$ exists,
is unique, and is differentiable. Now we notice that the objective function
of problem \eqref{eq:coll_prox_op} is
continuously-differentiable, and that $g(x,\af) \equiv \| \af (\ux - \ux') + (1 - \af) (\ox - \ox') \|^2 - (r + r')^2$
is continuously-differentiable in $(x,\af)$ and convex in $\af$. If it is true that 
$\partial_1 g(x^*(\af^*),\af^*) \neq 0$ then
we can apply Lemma \ref{th:lemma_for_max_min} with $\mathcal{A} = [0,1]$
and conclude that $x^*(\af^*)$ is also a solution of
problem \eqref{eq:coll_prox_op}. Hence the theorem will be true in this case as well.

% extra check
We now show that in this case, indeed, $\partial_1 g(x^*(\af^*),\af^*) \neq 0$.
First we notice that $\partial_1 g = (\partial_{\ux} g,\partial_{\ox} g,\partial_{\ux'} g,\partial_{\ox'} g) = 2 \| \af^* (\ux^* - {\ux^*}') + (1 - \af^*) (\ox^* - {\ox^*}') \| (\af^*, 1-\af^*, - \af^*, -1+\af^*)$. We now recall that, as explain above,
if $\| \af^* (\un - \un') + (1 - \af^*) (\on - \on') \| \neq 0$ then
$h(\af^*) > 0$. Therefore, we conclude that $r + r' =  \|\af^* (\ux^* - {\ux^*}') + (1 - \af^*) (\ox^* - {\ox^*}')\|$ because
otherwise $\|\af^* (\ux^* - {\ux^*}') + (1 - \af^*) (\ox^* - {\ox^*}')\| < r + r'$ implies that the constraint of problem
\eqref{eq:coll_prox_op_single_alpha} for $\af = \af^*$ is inactive which implies that the solution must be $x = n$ with objective value $0$ which contradicts the fact that $h(\af^*) > 0$.
Hence, $\partial_1 g = (r + r') (\af^*, 1-\af^*, - \af^*, -1+\af^*)$, which is always non-zero, and proves
that the theorem is true in this case as well.

% final proof
Finally, we now prove that that expressions \eqref{eq:expression_for_h}
to \eqref{eq:x_out_coll_4} hold when $\| \af (\un - \un') + (1 - \af) (\on - \on') \| \neq 0$.
This amounts to a relatively long calculus computation and is written in Appendix \ref{app:solving_single_alf_problem}.

\section{Computation of minimum value and minimizer of
problem \eqref{eq:coll_prox_op_single_alpha}} \label{app:solving_single_alf_problem}

We do not solve problem \eqref{eq:coll_prox_op_single_alpha} but instead solve the equivalent
(more smooth) problem
\begin{align} \label{eq:coll_prox_op_single_alpha_modified_constraint}
& {\min_{\ux,\ux',\ox,\ox'}} \urhof \| \ux - \un \|^2  + \orhof \| \ox - \on \|^2 \\
&+ \urhopf \| \ux' - \un' \|^2 + \orhopf \| \ox' - \on' \|^2 \nonumber\\
& \text{ s.t. } \| \af (\ux - \ux') + (1 - \af) (\ox - \ox') \|^2  - (r + r')^2 \geq 0 \nonumber.
\end{align}

To begin, we notice that if $$\| \af (\un - \un') + (1 - \af) (\on - \on') \| > (r + r')$$ then the 
constraint is inactive and the minimizer
of \eqref{eq:coll_prox_op_single_alpha_modified_constraint} is $(\un,\un',\on,\on')$
with minimum value $0$. 
At the same time, if $\| \af (\un - \un') + (1 - \af) (\on - \on') \| > (r + r')$ we have that $h(\af) = 0$
and the minimizer obtained from \eqref{eq:x_out_coll_1}-\eqref{eq:x_out_coll_4} is also
$(\un,\un',\on,\on')$. Therefore, we only need to show that equations \eqref{eq:expression_for_h} and \eqref{eq:x_out_coll_1}-\eqref{eq:x_out_coll_4}
hold in the case when the constraint is active, which corresponds to the case when 
$\| \af (\un - \un') + (1 - \af) (\on - \on') \| \leq (r + r')$.

To do so, we first introduce a few block variables (written in boldface) and express the above problem in a
shorter form. Namely, we define ${\bf x} = (\ux,\ux',\ox,\ox') \in \mathbb{R}^{4 \times d}$, ${\bf n} = (\un,\un',\on,\on')\in \mathbb{R}^{4 \times d}$ and ${\boldsymbol \af} = (\af , -\af  , (1-\af) , -(1-\af)) \in \mathbb{R}^{4}$ and $D = \text{diag}(\urho  ,\urho' ,\orho ,\orho' ) \in \mathbb{R}^{4 \times 4}$, and, rewrite \eqref{eq:coll_prox_op_single_alpha_modified_constraint} as
\begin{align} \label{eq:coll_prox_op_single_alpha_modified_constraint_short_form}
\begin{aligned}
&  &{\min_{\bf x}} \;\;\;
& \frac{1}{2} \text{tr} \{({\bf x - n})^{\dagger} D ({\bf x - n})\} \\
& & \text{s.t.}  \;\;\;
& \| {\boldsymbol \af}^{\dagger} {\bf x} \|^2  - (r + r')^2 \geq 0.
\end{aligned}
\end{align}
Then we notice that it is necessary that the solutions to this problem are among the
points that satisfy the KKT conditions. Namely, those points that satisfy
\begin{align}
&D ({\bf x - n}) + 2 {\boldsymbol \af} v= 0\\
&v = \lambda ({\boldsymbol \af}^{\dagger} {\bf x})\\
& \| v\|/|\lambda = r + r'
\end{align}
where $\lambda \neq 0$ is the Lagrange multiplier associated to the problem's constraint
and is non-zero because we are assuming the constraint is active.
In the rest of the proof we show that there are only two points that satisfy the KKT
conditions and show that, between them, the one that corresponds to the global
optimum satisfies \eqref{eq:expression_for_h} and \eqref{eq:x_out_coll_1}-\eqref{eq:x_out_coll_4}.

We first write the two equations even more compactly as
\begin{equation} \label{eq:compact_kkt_for_proof_of_main_theorem}
\left(
\begin{array}{cc}
\frac{1}{2} D & {\boldsymbol \af} \\
 {\boldsymbol \af}^{\dagger} & -1/\lambda
\end{array}\right)
\left(
\begin{array}{c}
{\bf x} \\
v
\end{array}\right)
= 
\left(
\begin{array}{c}
\frac{1}{2}D {\bf n} \\
0
\end{array}\right).
\end{equation}
We claim that, if $1+2\lambda {\boldsymbol \af}^\dagger D^{-1} {\boldsymbol \af} \neq 0$, the inverse of the block matrix
\begin{equation} \label{eq:main_matrix_to_prove_th_1}
\left(
\begin{array}{cc}
\frac{1}{2} D & {\boldsymbol \af} \\
 {\boldsymbol \af}^{\dagger} & -1/\lambda
\end{array}\right)
\end{equation}
is
\begin{equation} \label{eq:inverse_of_main_matrix_to_prove_th_1}
\left(
\begin{array}{cc}
2 \left(D^{-1}  - \frac{2 D^{-1} \lambda {\boldsymbol \af} {\boldsymbol \af}^\dagger D^{-1}}{1+2\lambda {\boldsymbol \af}^\dagger D^{-1} {\boldsymbol \af}} \right) & \frac{2 \lambda D^{-1} {\boldsymbol \af}}{1+2\lambda {\boldsymbol \af}^\dagger D^{-1} {\boldsymbol \af}} \\
\frac{2 \lambda {\boldsymbol \af}^\dagger D^{-1} }{1+2\lambda {\boldsymbol \af}^\dagger D^{-1} {\boldsymbol \af}} & \frac{-\lambda}{1+2\lambda {\boldsymbol \af}^\dagger D^{-1} {\boldsymbol \af}}
\end{array}\right).
\end{equation}
To prove this we could use the formula for the inverse of a block matrix. Instead, and much more simply,
we simply compute the product of \eqref{eq:inverse_of_main_matrix_to_prove_th_1} and \eqref{eq:main_matrix_to_prove_th_1} and show it equals the identify. It is immediate to see that
the block diagonal entries of the resulting product are indeed identity matrices. Since both matrices
are symmetric, all that is left to check is that one of the non-diagonal block entries is zero. Indeed,
{\small
\begin{align*}
&({\boldsymbol \af}^{\dagger}) \left(2 \left(D^{-1}  - \frac{2 D^{-1} \lambda {\boldsymbol \af} {\boldsymbol \af}^\dagger D^{-1}}{1+2\lambda {\boldsymbol \af}^\dagger D^{-1} {\boldsymbol \af}} \right) \right)\\
&+ (-1/\lambda) \left(\frac{2 \lambda {\boldsymbol \af}^\dagger D^{-1} }{1+2\lambda {\boldsymbol \af}^\dagger D^{-1} {\boldsymbol \af}}\right)\\
&= \frac{ 2 {\boldsymbol \af}^{\dagger} D^{-1} (1+2\lambda {\boldsymbol \af}^\dagger D^{-1} {\boldsymbol \af}) -4 {\boldsymbol \af}^{\dagger} D^{-1} \lambda {\boldsymbol \af} {\boldsymbol \af}^\dagger D^{-1} - 2 {\boldsymbol \af}^\dagger D^{-1}}{1+2\lambda {\boldsymbol \af}^\dagger D^{-1} {\boldsymbol \af}}\\
&=\frac{ 4\lambda  {\boldsymbol \af}^\dagger D^{-1} {\boldsymbol \af}^\dagger D^{-1} {\boldsymbol \af} -4 {\boldsymbol \af}^{\dagger} D^{-1} \lambda {\boldsymbol \af} {\boldsymbol \af}^\dagger D^{-1}}{1+2\lambda {\boldsymbol \af}^\dagger D^{-1} {\boldsymbol \af}}\\
& =
\frac{ ({\boldsymbol \af}^\dagger D^{-1} {\boldsymbol \af}) (4\lambda  {\boldsymbol \af}^\dagger D^{-1} -4  \lambda{\boldsymbol \af}^\dagger D^{-1})}{1+2\lambda {\boldsymbol \af}^\dagger D^{-1} {\boldsymbol \af}} = 0.
\end{align*}
}
We now solve the linear system \eqref{eq:compact_kkt_for_proof_of_main_theorem}
by multiplying both sides by the inverse matrix
\eqref{eq:inverse_of_main_matrix_to_prove_th_1} and, we conclude that 
\begin{align}
&v = \frac{2 \lambda {\boldsymbol \af}^\dagger D^{-1}}{1+2\lambda {\boldsymbol \af}^\dagger D^{-1} {\boldsymbol \af}} \left(\frac{1}{2} D {\bf n}\right) = \frac{\lambda {\boldsymbol \af}^\dagger {\bf n}}{1+2\lambda {\boldsymbol \af}^\dagger D^{-1} {\boldsymbol \af}},\label{eq:almost_ready_expression_for_v}\\
&{\bf x} = 2 \left(D^{-1}  - \frac{2 D^{-1} \lambda {\boldsymbol \af} {\boldsymbol \af}^\dagger D^{-1}}{1+2\lambda {\boldsymbol \af}^\dagger D^{-1} {\boldsymbol \af}} \right) \left(\frac{1}{2}D {\bf n} \right)\nonumber\\
& = {\bf n} - \frac{2 D^{-1} \lambda {\boldsymbol \af} {\boldsymbol \af}^\dagger {\bf n}}{1+2\lambda {\boldsymbol \af}^\dagger D^{-1} {\boldsymbol \af}}  \label{eq:almost_ready_expression_for_x}.
\end{align}

Using equation \eqref{eq:almost_ready_expression_for_x} we can express the objective value
of \eqref{eq:coll_prox_op_single_alpha_modified_constraint_short_form} as
\begin{align*}
&\frac{1}{2} \text{tr} \{ ({\bf x - n})^{\dagger} D ({\bf x - n}) \}\\
&= \frac{\text{tr} \{ {\bf n}^{\dagger} {\boldsymbol \af} {\boldsymbol \af}^{\dagger} \lambda D^{-1} (-2) D (-2) D^{-1} \lambda {\boldsymbol \af} {\boldsymbol \af}^{\dagger} {\bf n} \}}{2(1+2\lambda {\boldsymbol \af}^\dagger D^{-1} {\boldsymbol \af})^2} \\
&=\frac{\text{tr} \{ {\bf n}^{\dagger} {\boldsymbol \af} {\boldsymbol \af}^{\dagger} {\bf n} \} 4 \lambda^2  ({\boldsymbol \af}^{\dagger} D^{-1} {\boldsymbol \af}) }{2(1+2\lambda {\boldsymbol \af}^\dagger D^{-1} {\boldsymbol \af})^2}\\
&=\frac{ 2 \lambda^2 \| {\boldsymbol \af}^{\dagger} {\bf n} \|^2 ({\boldsymbol \af}^{\dagger} D^{-1} {\boldsymbol \af}) }{(1+2\lambda {\boldsymbol \af}^\dagger D^{-1} {\boldsymbol \af})^2}
=  2 \|v\|^2 ({\boldsymbol \af}^{\dagger} D^{-1} {\boldsymbol \af}),
\end{align*}
where in the last equality we made use of \eqref{eq:almost_ready_expression_for_v}.
We now recall that from the third equation in the KKT conditions we have
that $\|v\|/\lambda = r + r'$ and so we conclude that 
\begin{equation} \label{eq:two_expressions_for_lambda}
\lambda = \frac{1}{2 ( {\boldsymbol \af}^\dagger D^{-1} {\boldsymbol \af})}\left(-1 \pm \frac{\|{\boldsymbol \af}^\dagger {\bf n}\|}{r + r'} \right),
\end{equation}
and therefore,
\begin{align}
&\frac{1}{2} \text{tr} \{ ({\bf x - n})^{\dagger} D ({\bf x - n}) \}\nonumber\\
&=
 2 (r + r')^2 \frac{1}{4 ({\boldsymbol \af}^{\dagger} D^{-1} {\boldsymbol \af})^2} \left(-1 \pm \frac{\|{\boldsymbol \af}^\dagger {\bf n}\|}{r + r'} \right)^2 ({\boldsymbol \af}^{\dagger} D^{-1} {\boldsymbol \af})\nonumber\\
& =\frac{(r + r' \pm \| {\boldsymbol \af}^{\dagger} {\bf n}\|)^2}{2 ({\boldsymbol \af}^\dagger D^{-1} {\boldsymbol \af})}.
\end{align}
Since we are seeking the global minimum, and since we are assuming
that $\|{\boldsymbol \af}^{\dagger} {\bf n} \| = \| \af (\un - \un') + (1 - \af) (\on - \on') \| \leq (r + r')$ and hence the constraint is active, we conclude that
\begin{align}
&\frac{1}{2} \text{tr} \{ ({ x^*(\af) - n})^{\dagger} D ({x^*(
\af) - n}) \}\nonumber\\
&= \frac{(r + r' - \| {\boldsymbol \af}^{\dagger} {\bf n}\|)^2}{2 ({\boldsymbol \af}^\dagger D^{-1} {\boldsymbol \af})}= \frac{1}{2} (h(\af))^2.
\end{align}
Above we have used the fact that ${\boldsymbol \af}^{\dagger} {\bf n} = \af \Delta \un + (1 - \af) \Delta \on$ and that ${\boldsymbol \af}^\dagger D^{-1} {\boldsymbol \af} = \af^2 / \utilde{\rho} + (1- \af)^2 / \tilde{\rho}$. This proves that \eqref{eq:expression_for_h} is valid when
$\|{\boldsymbol \af}^{\dagger} {\bf n} \| = \| \af (\un - \un') + (1 - \af) (\on - \on') \| \leq (r + r')$
as long as $1+2\lambda {\boldsymbol \af}^\dagger D^{-1} {\boldsymbol \af} = \|{\boldsymbol \af}^{\dagger} {\bf n} \|/(r + r') \neq 0$.
Recall that we have already proved that \eqref{eq:expression_for_h} holds when 
when $\|{\boldsymbol \af}^{\dagger} {\bf n} \| = \| \af (\un - \un') + (1 - \af) (\on - \on') \| > (r + r')$.

To finish the proof we now prove the validity of \eqref{eq:x_out_coll_1}-\eqref{eq:x_out_coll_4} 
when $\|{\boldsymbol \af}^{\dagger} {\bf n} \| = \| \af (\un - \un') + (1 - \af) (\on - \on') \| > (r + r')$.
Recall that we have already proved their validity when $\|{\boldsymbol \af}^{\dagger} {\bf n} \| = \| \af (\un - \un') + (1 - \af) (\on - \on') \| \leq (r + r')$.
Now notice that, when $\|{\boldsymbol \af}^{\dagger} {\bf n} \| = \| \af (\un - \un') + (1 - \af) (\on - \on') \| > (r + r')$, we can write,
\begin{align}
&\lambda = \frac{1}{2 ( {\boldsymbol \af}^\dagger D^{-1} {\boldsymbol \af})} \left( -1 + \frac{\|{\boldsymbol \af}^\dagger {\bf n}\|}{r + r'} \right)\nonumber\\
&= - \frac{h(\af)}{2 (r +r') \sqrt{{\boldsymbol \af}^\dagger D^{-1} {\boldsymbol \af}}}.
\end{align}
If we define $\gamma  = \frac{2 \lambda}{1 + 2 \lambda {\boldsymbol \af}^\dagger D^{-1} {\boldsymbol \af}}$,
we can write the following expression for ${ x}^*(\af)$
\begin{equation}
{ x}^*(\af) = {\bf n} - \gamma D^{-1}  {\boldsymbol \af}  {\boldsymbol \af}^{\dagger} {\bf n},
\end{equation}
which holds if $\|{\boldsymbol \af}^{\dagger} {\bf n} \| = \| \af (\un - \un') + (1 - \af) (\on - \on') \| \neq  0$.
With a little bit of algebra one can see that this is exactly the same expression as
\eqref{eq:x_out_coll_1}-\eqref{eq:x_out_coll_4} and finish the proof.

\end{document}